\pdfoutput=1
\def\year{2021}\relax
\documentclass[letterpaper]{article} 
\usepackage{aaai21}  
\usepackage{times}  
\usepackage{helvet} 

\usepackage{courier}  
\usepackage[hyphens]{url}  
\usepackage{graphicx} 
\usepackage{natbib}  
\usepackage{caption} 
\title{Raven's Progressive Matrices Completion with Latent Gaussian Process Priors}
\author{
    Fan Shi, Bin Li\protect\thanks{Corresponding author}, Xiangyang Xue
    \\
}
\affiliations{
 Shanghai Key Laboratory of Intelligent Information Processing\\
 School of Computer Science, Fudan University\\
 \{fshi19, libin, xyxue\}@fudan.edu.cn
}

\usepackage{amsmath}
\usepackage{amsfonts, amssymb}
\usepackage[switch]{lineno}

\begin{document}

\maketitle

\begin{abstract}
Abstract reasoning ability is fundamental to human intelligence. It enables humans to uncover relations among abstract concepts and further deduce implicit rules from the relations. As a well-known abstract visual reasoning task, Raven's Progressive Matrices (RPM) are widely used in human IQ tests. Although extensive research has been conducted on RPM solvers with machine intelligence, few studies have considered further advancing the standard answer-selection (classification) problem to a more challenging answer-painting (generating) problem, which can verify whether the model has indeed understood the implicit rules. In this paper we aim to solve the latter one by proposing a deep latent variable model, in which multiple Gaussian processes are employed as priors of latent variables to separately learn underlying abstract concepts from RPMs; thus the proposed model is interpretable in terms of concept-specific latent variables. The latent Gaussian process also provides an effective way of extrapolation for answer painting based on the learned concept-changing rules. We evaluate the proposed model on RPM-like datasets with multiple continuously-changing visual concepts. Experimental results demonstrate that our model requires only few training samples to paint high-quality answers, generate novel RPM panels, and achieve interpretability through concept-specific latent variables.
\end{abstract}

\begin{figure}[t]
\centering
\includegraphics[width=0.8\columnwidth]{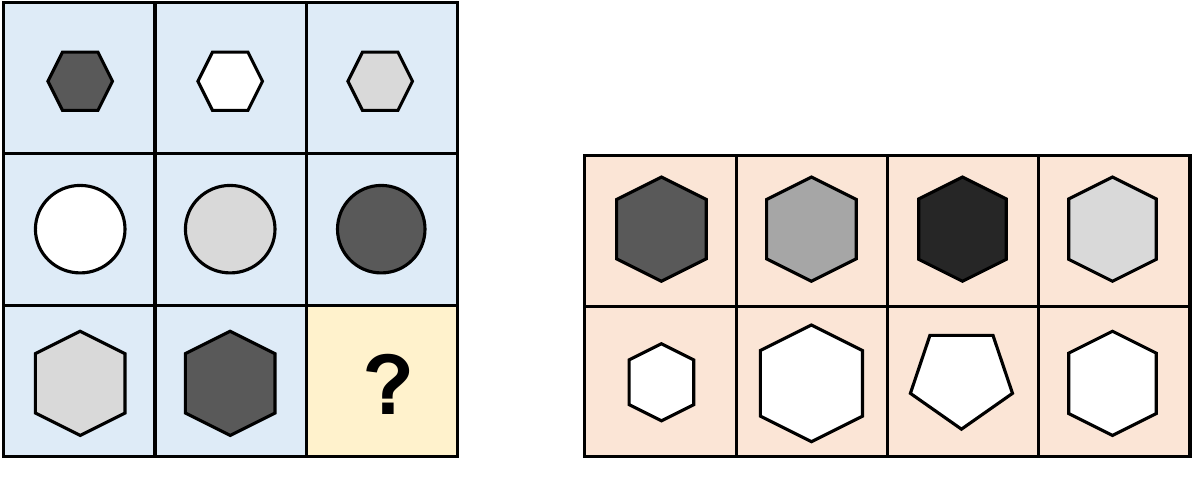}
\caption{An example of the RPM panel from the RAVEN dataset \cite{zhang2019raven}.
The context panel (left) consists of eight observed cells in blue background and one missing cell in yellow background; the selection panel (right) consists of eight cells in orange background.}
\label{fig:rpm_samples}
\end{figure}

\noindent A major performance of human abstract thinking is the ability to discover relations between abstract concepts and then summarize underlying rules. Abstract visual reasoning tasks reflect such abstract thinking capabilities. The Raven's Progressive Matrices (RPM) problem, a kind of nonverbal abstract visual reasoning task proposed by John C. Raven \cite{raven1938raven}, are usually used to test human fluid intelligence \cite{bilker2012development}. As an example shown in Figure \ref{fig:rpm_samples}, participants of the RPM problem should observe a $3 \times 3$ context panel with the right-bottom cell invisible. They are expected to find out underlying concept-changing rules and choose an answer \cite{raven1938raven} from a selection panel. The RPM problem predominantly emphasizes two aspects of intelligence: visual representation and abstract reasoning. From the perspective of visual representation, our vision and nervous system tend to analyze independent visual concepts from an observed object: such as $color$, $shape$, and $size$. From the perspective of abstract reasoning, when conducting RPM IQ tests, our strong abstract reasoning ability is based on the comprehension of independent visual concepts and also, concept-changing rules.

In the field of machine learning, enabling computational models with the abstract reasoning ability is challenging. Some similarity-based models \cite{barrett2018measuring} measure similarity between every two cell content to carry out the reasoning process. Recently, as researchers show more interest in deep learning, a considerable amount of literature has been published on methods utilizing deep neural networks \cite{barrett2018measuring,zhang2019learning,zhang2019raven}. For a particular downstream task, the choice of feature representation has a great impact on performance. If feature representations are well adapted to the current task, the accuracy and robustness of models can be improved \cite{bengio2013representation}. The variational autoencoder (VAE) \cite{kingma2013auto} has emerged as a functional tool in the unsupervised representation learning. To enhance the interpretability in representations, the notion of disentanglement is put forward. By decomposing and regularizing the evidence lower bound (ELBO) of VAEs, a number of disentangled latent variable models have been proposed \cite{higgins2016beta,kumar2017variational,kim2018disentangling,chen2018isolating}. These regularization-based models force each dimension (or multiple non-overlapping dimensions) of latent variables to code a single visual concept, which endows these dimensions clear meanings. Some recent studies \cite{steenbrugge2018improving,van2019disentangled} point out that replacing deep representations in the Wild Relational Network (WReN) \cite{barrett2018measuring} with disentangled representations will improve the selection accuracy and data efficiency of the model when solving the RPM problem.

The related studies indicate substantial progress in the RPM problem. However, these methods still have downsides: (1) Similarity-based RPM solvers fail to capture concept-changing rules and lack the explicit generative process of entire panels; (2) Methods that introduce representations from the regularization-based disentangled VAE struggle to encode reliable concept-specific latent variables \cite{locatello2019challenging}. Both drawbacks spoil the interpretability of models. In fact, human cognition and abstract reasoning abilities are not limited in the original selective RPM problem. While tested with easily recognized rules, a human can paint the correct answer after inspecting context panels. Most RPM solvers nowadays focus on the regular cell selection problem. However, the entire panel completion is a more challenging task that further evaluates the abstract reasoning ability to find concept-changing rules.

In this paper, we propose an interpretable deep latent variable model that explicitly defines the generative and inference process of entire panels as well as regards the prior of concept-changing rules as 2D Gaussian processes (GPs) with the learnable kernel \cite{wilson2016deep} to solve panel completing tasks. Structure information within context panels is introduced to guide the learning of concept-specific representations. Due to the continuity of GPs and the concept-oriented characteristic of representations, it is possible to extrapolate answers from independent visual concepts of the observed cell contents. Deduced answer representations and context cell representations are finally decoded to construct complete panel images.

We evaluate the model in RPM-like datasets with continuously changing visual concepts. To evaluate both the reconstruction quality and comprehension of underlying concept-changing rules, we calculate the averaged MSE over panel cells by leaving out and predicting one cell in turn.  For all datasets, the proposed model outperforms the VAE-GAN inpainting model \cite{yu2018generative}. In addition to the missing cell prediction, our model can also generate novel panels according to deduced concept-changing rules, which has not been investigated in previous works. Further experiments about the multi-cell prediction and visual concept visualization indicate that our model realizes interpretability by means of the concept-specific latent variables and human-like rule reasoning process.

\section{Related Work}

\subsection{RPM Problem Solver}

To solve the RPM problem, some computational models are proposed \cite{lovett2017modeling,little2012bayesian,lovett2010structure} to extract symbolic representations from the context panel and perform inference based on the particular rules. With the extensive application of deep neural networks, more and more models use convolutional neural networks as image feature extractors. A novel method called WReN \cite{barrett2018measuring}, based on the Relation Network \cite{santoro2017simple} and similarity scores, achieves competitive performance in the RPM problem. To evaluate the WReN, an RPM-like dataset named Procedurally Generated Matrices (PGM) is built. CoPINet \cite{zhang2019learning} introduces the contrast effect from humans and animals to enhance the reasoning ability. In \cite{zhang2019raven}, authors release a more complex dataset RAVEN and use a tree-structured neural module called Dynamic Residual Tree (DRT) to represent the intrinsic structure of images. However, there exist obvious patterns in RAVEN's selection panels, which leads to the high classification accuracy even without observing context panels \cite{hu2020hierarchical}. To overcome this defect, an unbiased Balanced-RAVEN dataset is proposed \cite{hu2020hierarchical}. Then the V-PROM dataset \cite{teney2020v} extends the image style inside cells to more complex visual scenes. On the other hand, some studies on the visual representation \cite{steenbrugge2018improving,van2019disentangled} reveal that disentangled representations can significantly improve model performance in the RPM problem.

\subsection{Disentangled VAE}

The disentangled VAE provides an idea for the concept-specific representation learning.

Regularization-based methods impose regularization constraints to the posterior distribution or aggregate posterior distribution in ELBO. $\beta$-VAE \cite{higgins2016beta} reformulates maximization of the objective function to an optimization problem with inequality constraints. While reconstructing the real data, these constraints force the posterior distribution closed to an isotropic standard Gaussian prior. DIP-VAE \cite{kumar2017variational} adopts an extra regularization term to reduce the distance between the aggregate posterior and prior by moment matching. FactorVAE \cite{kim2018disentangling} and $\beta$-TCVAE \cite{chen2018isolating} deal with the mutual information over-penalization problem in $\beta$-VAE. Both methods further decompose the ELBO and exactly exert a regularization penalty in the total correlation to ensure the independence between dimensions. $\beta$-TCVAE estimates the total correlation by minibatch-weighted sampling, while FactorVAE uses a density ratio trick that requires an additional discriminator.

In \cite{locatello2019challenging}, the authors point out the importance of explicit inductive bias and supervision information in the disentangled representation learning. By contrast, some methods introduce inductive bias, such as distinctive structures in data and network architectures, into models. VITAE \cite{skafte2019explicit} combines a symmetric transformation and an inverse transformation operations in the computation process to disentangle the appearance and perspective of objects. CascadeVAE \cite{jeong2019learning} and JointVAE \cite{dupont2018learning} assume that the latent space consists of discrete and continuous parts and disentangle both discrete and continuous representations. Spatial Broadcast Decoder \cite{watters2019spatial} broadcasts and concatenates latent variables to improve the disentanglement ability.

\subsection{GP with Deep Kernels}

A GP is a set of random variables whose finite subsets satisfy the multivariate Gaussian distribution \cite{williams2006gaussian}, which can be expressed as $f(\textbf{x}) \sim \mathcal{G} \mathcal{P}(m(\textbf{x}), \textbf{K}_{XX})$ where $m(\textbf{x})$ is the mean function usually set to zero function and $\textbf{K}_{XX}$ denotes the covariance matrix calculated from predefined kernel functions.

To improve the flexibility and avoid the empirical selection of kernels, some studies \cite{wilson2016deep,sun2018differentiable} incorporate the GP kernel with deep neural networks. In the function-space view, a GP defines the particular distribution over functions whose basic characters (e.g., smoothness) are determined via kernels. Therefore, constructing or choosing kernels is an essential work in the GP prediction. In \cite{wilson2016deep}, the model encodes original locations with a deep structure consisting of nonlinear mappings whose output is used in the spectral mixture base kernel. Neural Kernel Network \cite{sun2018differentiable} exploits composition rules to form more complicated kernels by considering each unit of a neural network as one basic kernel. Most works attempt to use the GP with deep kernels to create function spaces mapping data to predictive values. However, in this paper, we regard deep kernel GPs as the prior of concept-changing rules whose input and output reside in latent variable spaces.

\section{The Proposed Method}

Inspired by the discussion in previous works about disentangled representations in the original RPM problem \cite{van2019disentangled}, we propose a deep latent variable model to solve the RPM painting problem. The model firstly encodes context cell images (objects inside the panel cells) to independent concept-specific latent variables. The latent variables of missing cell images are then derived via the GP rule reasoning where GPs are taken as the priors of underlying concept-changing rules. Finally, all latent variables are concatenated and sent to a decoder to reconstruct the original context panels as well as missing cell images. In this section, we will introduce the GP rule reasoning, generative model, inference model, and parameter learning method in detail.

\begin{figure}[t]
\centering
\includegraphics[width=0.9\columnwidth]{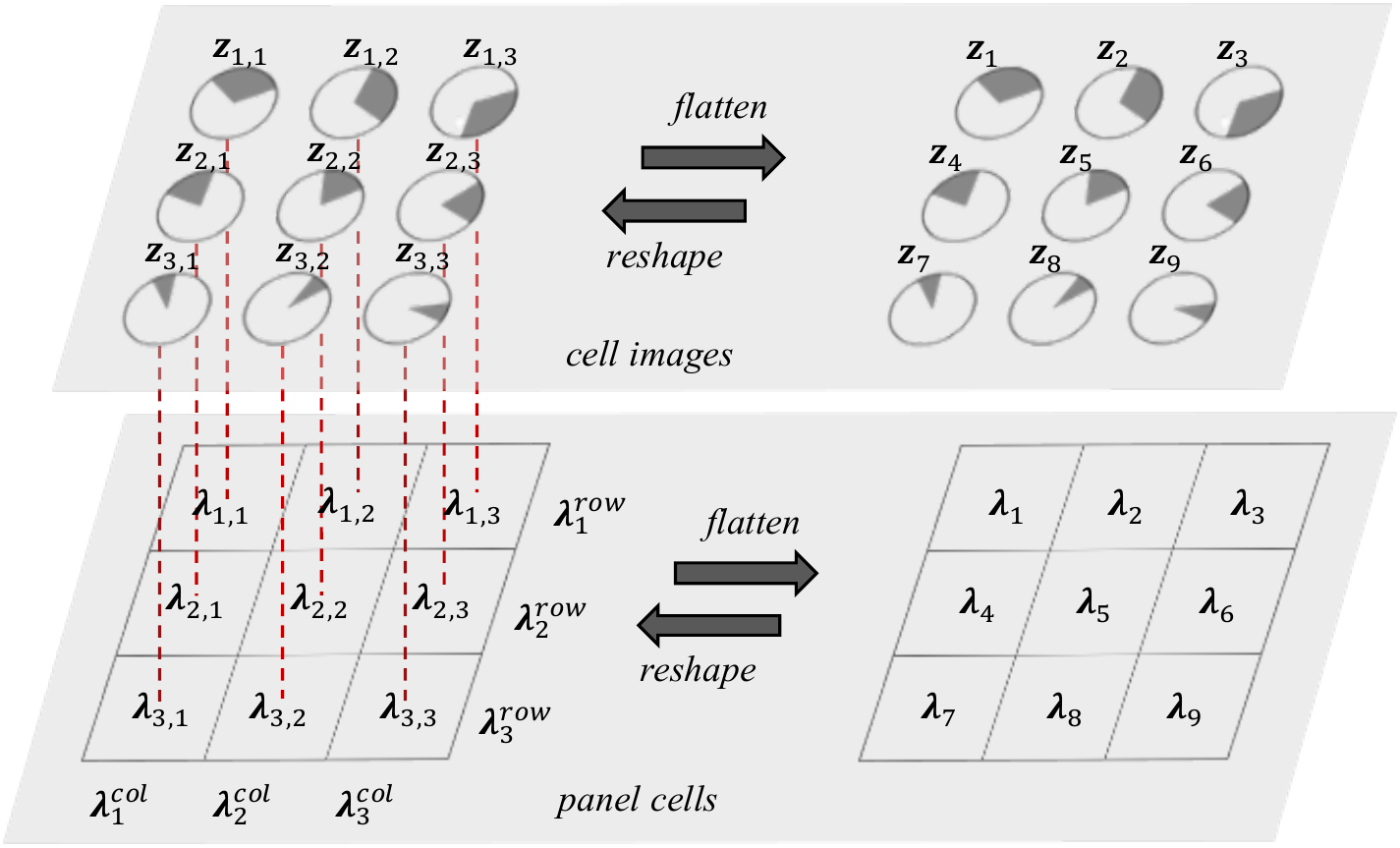}
\caption{An illustration of the RPM panel and its notations.}
\label{fig:panel_notations}
\end{figure}

\subsection{GP Rule Reasoning}

In Figure \ref{fig:panel_notations}, a complete RPM panel contains 3 rows and columns. We use $\boldsymbol{\lambda}^{row}=\{\boldsymbol{\lambda}_{1}^{row},\boldsymbol{\lambda}_{2}^{row},\boldsymbol{\lambda}_{3}^{row}\}$ and $\boldsymbol{\lambda}^{col}=\{\boldsymbol{\lambda}_{1}^{col},\boldsymbol{\lambda}_{2}^{col},\boldsymbol{\lambda}_{3}^{col}\}$ to denote row and column latent variables. By combining $\boldsymbol{\lambda}^{row}$ and $\boldsymbol{\lambda}^{col}$, we can represent the panel cell in the $i$th row and $j$th column by a tuple $(\boldsymbol{\lambda}_{i}^{row},\boldsymbol{\lambda}_{j}^{col})$. For convenience of notations, we define that $\boldsymbol{\lambda}_{i,j} = (\boldsymbol{\lambda}_{i}^{row},\boldsymbol{\lambda}_{j}^{col})$. Then we define panel's underlying concept-changing rules as 2D functions $f^{d}: R \times R \rightarrow R$ that map $\boldsymbol{\lambda}_{i,j}$ to cell image latent variables $\boldsymbol{z}_{i,j}$ and we have $f^{d}(\boldsymbol{\lambda}_{i,j}) = z_{i,j}^{d}$, where $i,j \in \{1, 2, 3\}$ and $d$ is the index of $\boldsymbol{z}_{i,j}$'s dimension varying from 1 to $D$. Considering that each dimension corresponds to a single visual concept, the proposed method tends to learn different concept-changing rules for every dimension. To introduce rule functions into the Bayesian learning framework, we intuitively choose GPs as the priors. Another index $k=1, ..., 9$ is used to denote the flattened 2D latent variables $\boldsymbol{\lambda}$ and $\boldsymbol{z}$. We have $k=3(i-1)+j$ and $i,j \in \{1, 2, 3\}$ (e.g., $\boldsymbol{z}_{6} = \boldsymbol{z}_{2,3}$ in Figure \ref{fig:panel_notations}). The $d$th dimension of $\boldsymbol{z}$ satisfies a multivariate Gaussian distribution.
\begin{equation}
\begin{split}
    \boldsymbol{\lambda}_{1:9} &= flatten\left(\boldsymbol{\lambda}\right) \\
    \boldsymbol{K}_{\boldsymbol{\theta}}^{d} &= \left[\begin{array}{ccc}
    \kappa_{\boldsymbol{\theta}}^{d}\left(\boldsymbol{\lambda}_{1}, \boldsymbol{\lambda}_{1}\right) & \cdots & \kappa_{\boldsymbol{\theta}}^{d}\left(\boldsymbol{\lambda}_{1}, \boldsymbol{\lambda}_{9}\right) \\
    \vdots & \ddots & \vdots \\
    \kappa_{\boldsymbol{\theta}}^{d}\left(\boldsymbol{\lambda}_{9}, \boldsymbol{\lambda}_{1}\right) & \cdots & \kappa_{\boldsymbol{\theta}}^{d}\left(\boldsymbol{\lambda}_{9}, \boldsymbol{\lambda}_{9}\right)
    \end{array}\right] \\
    \boldsymbol{z}_{1:9}^{d} &\sim \mathcal{N}\left(\boldsymbol{0}, \boldsymbol{K}_{\boldsymbol{\theta}}^{d}\right) \\
    \boldsymbol{z}^{d} &= reshape\left(\boldsymbol{z}_{1:9}^{d}\right)
\end{split}
\label{eq:draw-panel}
\end{equation}
where $\kappa_{\boldsymbol{\theta}}^{d}(., .)$ is the kernel function utilizing a neural network $g_{\boldsymbol{\theta}}^{d}$ and RBF kernel parameterized with $l$ and $\sigma$ as in \cite{wilson2016deep}.
\begin{equation}
    \kappa_{\boldsymbol{\theta}}^{d}\left(\boldsymbol{\lambda}_{m}, \boldsymbol{\lambda}_{n}\right) = l^{2} \exp\left(\frac{\left\|g_{\boldsymbol{\theta}}^{d}\left(\boldsymbol{\lambda}_{m}\right)-g_{\boldsymbol{\theta}}^{d}\left(\boldsymbol{\lambda}_{n}\right)\right\|_{2}^{2}}{2 \sigma^{2}}\right)
\end{equation}
It is the learnable parameters in $\kappa_{\boldsymbol{\theta}}^{d}(., .)$ that realize the automatic function space adaption for real data. $flatten$ denotes a function that flattens a matrix to a vector and $reshape$ denotes an inverse procedure that deforms a vector into a matrix. The complete process described in Eq.(\ref{eq:draw-panel}) is referred to as $\boldsymbol{z}^{d} = panelSampling(\boldsymbol{\lambda}, d)$. By means of the GP priors, we can perform the following operations readily.

\textbf{Complete Panels Sampling.} As described above, we can calculate kernels $\boldsymbol{K}_{\boldsymbol{\theta}}^{d}$ via $\boldsymbol{\lambda}$ and then separately sample $\boldsymbol{z}^{d}$ from dimension-wise multivariate Gaussian distributions defined in Eq.(\ref{eq:draw-panel}). In the remaining generative process, $\boldsymbol{z}^{d}$ are concatenated through dimensions and then decoded to form a complete panel.

\textbf{Conditional Cells Prediction.} One significant facility of GPs is the ability to predict function values at novel locations. In the same way, we can infer the dimension-wise answer latent variable $z_{3,3}^{d}$ based on the latent variables of the observed cell images $\boldsymbol{z}_{\neg 3,3}^{d}$, where the notation $\neg 3,3$ means leaving the element indexed $3,3$ out. This process gives the proposed method ability of abstract reasoning, which is utilized in the inference phase.

\subsection{Generative Model}

\begin{figure}[t]
\centering
\includegraphics[width=0.85\columnwidth]{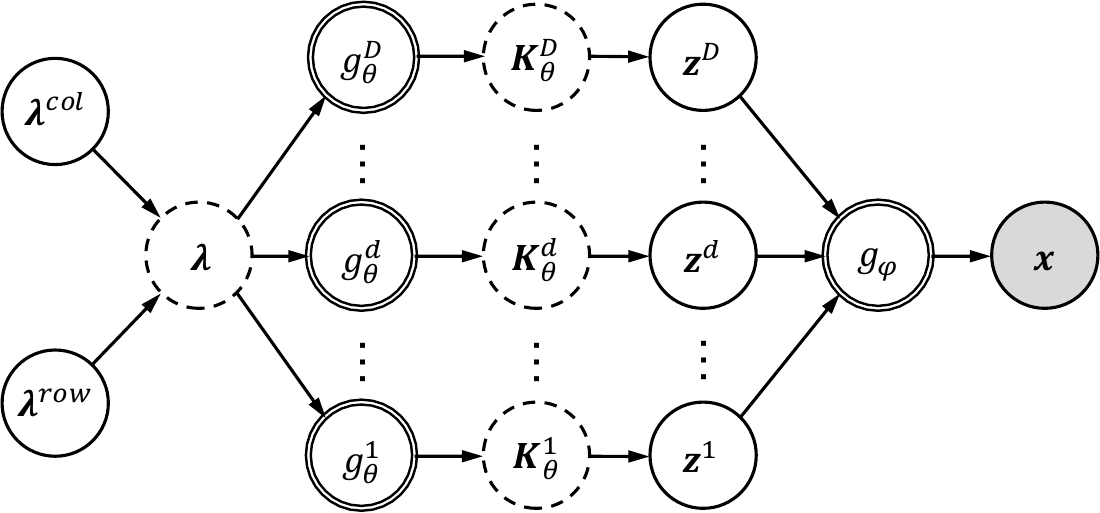}
\caption{The diagram of the generative model. Solid circles are random variables including both latent and observed variables. Double solid circles mean neural networks and dashed circles denote determinate variables.}
\label{fig:generative_model}
\end{figure}

According to Figure \ref{fig:generative_model}, the process to generate the cell images $\boldsymbol{x}$ of the entire panel from $\boldsymbol{z}$, $\boldsymbol{\lambda}^{row}$, and $\boldsymbol{\lambda}^{col}$ is
\begin{equation}
\begin{split}
    \boldsymbol{\lambda}^{row} &\sim \mathcal{N} \left(\boldsymbol{0}, \sigma_{\lambda}^{2}\boldsymbol{I} \right) \\
    \boldsymbol{\lambda}^{col} &\sim \mathcal{N} \left(\boldsymbol{0}, \sigma_{\lambda}^{2}\boldsymbol{I} \right) \\
    \boldsymbol{z}^{d} &= panelSampling\left( \boldsymbol{\lambda}, d\right), \quad d = 1, 2, ..., D \\
    \boldsymbol{z} &= concat\left(\boldsymbol{z}^{1}, ..., \boldsymbol{z}^{D}\right) \\
    \boldsymbol{x}_{i,j} &\sim \mathcal{N}\left(g_{\varphi}\left(\boldsymbol{z}_{i,j}\right), \sigma_{x}^{2}\boldsymbol{I}\right), \quad i,j \in \{1, 2, 3\}
\end{split}
\end{equation}
$\boldsymbol{\lambda}^{row}$ and $\boldsymbol{\lambda}^{col}$ are respectively panel's rows and columns that form the 2D locations $\boldsymbol{\lambda}$ of rule functions. Their priors are isotropic multivariate Gaussian distributions parameterized with $\sigma_{\lambda}$. In function $panelSampling$ (defined in the previous section GP Rule Reasoning), dimension-wise covariance matrices $\boldsymbol{K}_{\boldsymbol{\theta}}^{d}$ are calculated based on the learnable kernel functions. Then rule functions can be sampled from the GPs. Since no Gaussian noise is introduced to observations, $\boldsymbol{z}^{d}$ is equal to the function value sampled from the multivariate Gaussian distribution. Once $\boldsymbol{z}^{d}$ is stacked dimension by dimension, a neural network $g_{\boldsymbol{\varphi}}$ will decode $\boldsymbol{z}$ to means of the Gaussian distribution over $\boldsymbol{x}$. Variances of the Gaussian distribution over $\boldsymbol{x}$ are artificially determined by a constant $\sigma_{x}$. The joint probability of latent variables $\boldsymbol{\eta} = \{\boldsymbol{z}, \boldsymbol{\lambda}^{row}, \boldsymbol{\lambda}^{col}\}$, and $\boldsymbol{x}$ is
\begin{equation}
\begin{split}
    p_{\boldsymbol{\varphi}, \boldsymbol{\theta}}&\left(\boldsymbol{x},\boldsymbol{\eta}\right) = \prod_{i=1}^{3}\prod_{j=1}^{3}p_{\boldsymbol{\varphi}}\left(\boldsymbol{x}_{i,j}\mid \boldsymbol{z}_{i,j}\right)p\left(\boldsymbol{\lambda}^{row}\right) \\ &p\left(\boldsymbol{\lambda}^{col}\right)\prod_{d=1}^{D}p_{\boldsymbol{\theta}}\left(\boldsymbol{z}^{d}\mid \boldsymbol{\lambda}^{row},\boldsymbol{\lambda}^{col}\right)
\end{split}
\end{equation}

\subsection{Variational Inference}

\begin{figure}[t]
\centering
\includegraphics[width=0.7\columnwidth]{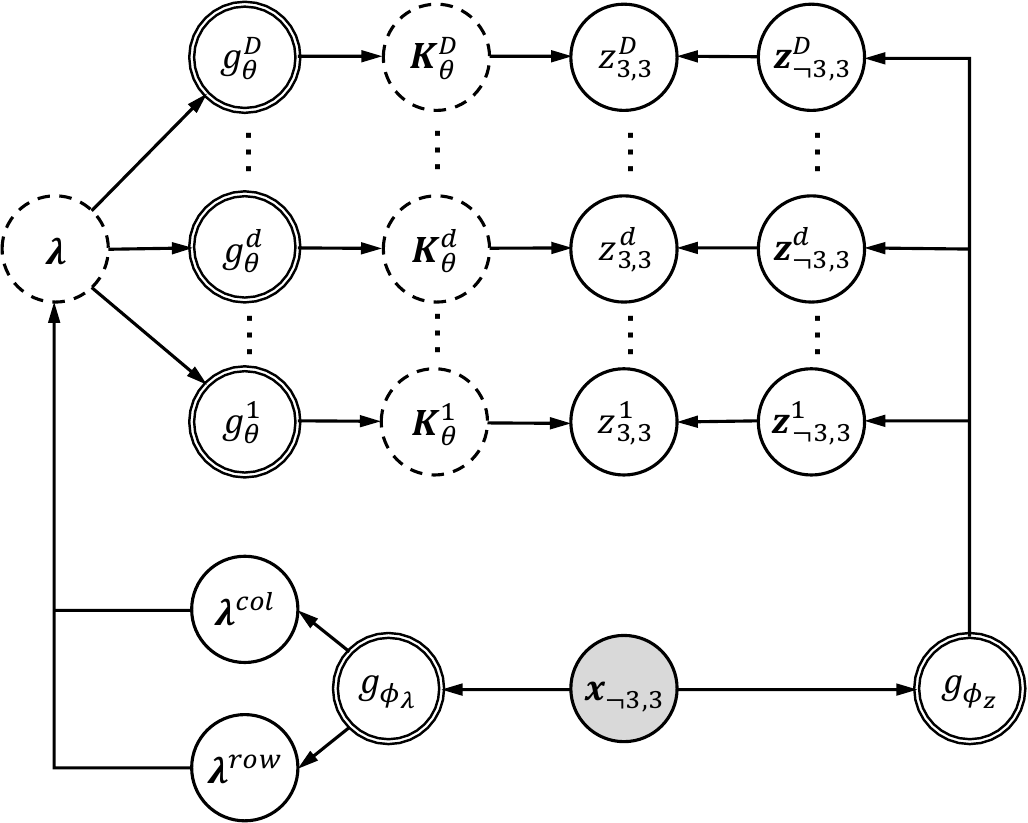}
\caption{The diagram of the inference model. The meaning of distinct circles are the same as those in Figure \ref{fig:generative_model}.}
\label{fig:inference_model}
\end{figure}

\begin{figure*}[t]
\centering
\includegraphics[width=0.95\textwidth]{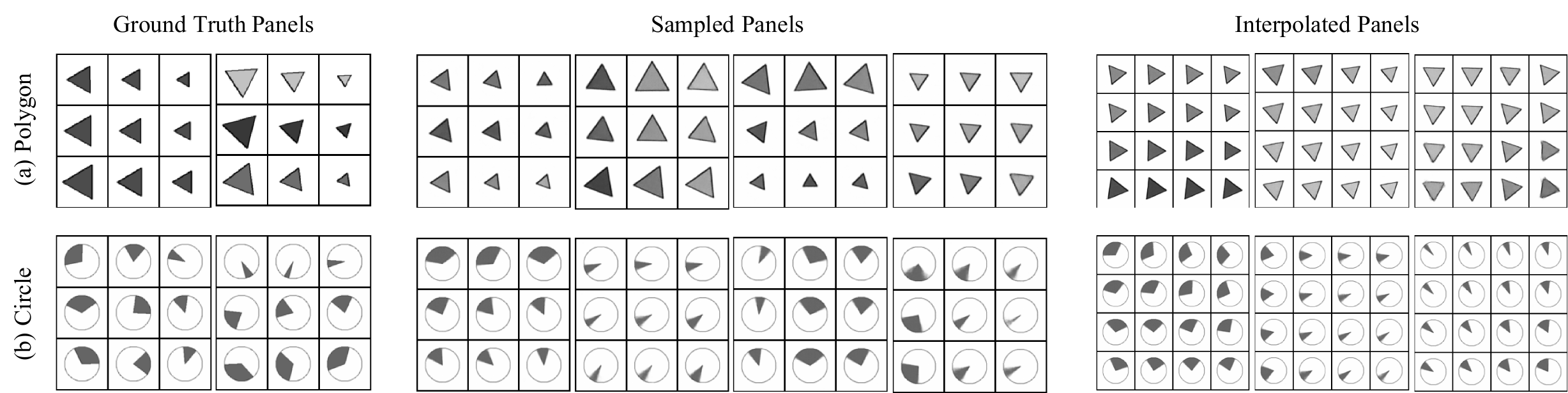} 
\caption{Panels sampled from datasets and the generative model. The left side are ground truth panels from datasets. Panels sampled from the generative model is displayed in the middle. The right side contains panels sampled via interpolation.}
\label{fig:exp_panel_generate}
\end{figure*}

In the framework of VAE, since the exact posterior distribution is hard to derive, a variational distribution is used to approximate the posterior distribution. Following the diagram in Figure \ref{fig:inference_model}, the variational posterior distribution can be expressed as
\begin{equation}
\begin{split}
    q_{\boldsymbol{\phi}, \boldsymbol{\theta}}&\left(\boldsymbol{\eta}\mid\boldsymbol{x}_{\neg 3,3}\right) = \prod_{d=1}^{D}q_{\boldsymbol{\theta}}\left(\boldsymbol{z}_{3,3}^{d}\mid\boldsymbol{z}_{\neg 3,3}^{d},\boldsymbol{\lambda}^{row},\boldsymbol{\lambda}^{col}\right)\\
    &q_{\boldsymbol{\phi_{\lambda}}}\left(\boldsymbol{\lambda}^{row}\mid\boldsymbol{x}_{\neg 3,3}\right)
    q_{\boldsymbol{\phi_{\lambda}}}\left(\boldsymbol{\lambda}^{col}\mid\boldsymbol{x}_{\neg 3,3}\right) \\
    &\prod_{(i,j) \in \Omega}q_{\boldsymbol{\phi_{z}}}\left(\boldsymbol{z}_{i,j}\mid\boldsymbol{x}_{i,j}\right)
\end{split}
\end{equation}
where $\boldsymbol{\phi} = \left\{\boldsymbol{\phi_{\lambda}}, \boldsymbol{\phi_{z}} \right\}$ denote the neural network $g_{\boldsymbol{\phi_{\lambda}}}$ and $g_{\boldsymbol{\phi_{z}}}$, respectively; $\Omega=\{1,2,3\}^{2}\setminus(3,3)$ is the set for the index $(i,j)$. Variational distributions $q_{\boldsymbol{\phi_{\lambda}}}(\boldsymbol{\lambda}^{row}|\boldsymbol{x}_{\neg 3,3})$ and $q_{\boldsymbol{\phi_{\lambda}}}(\boldsymbol{\lambda}^{col}|\boldsymbol{x}_{\neg 3,3})$ are both Gaussian distributions denoting rows and columns of the context panel. Means and standard deviations of the distribution are outputs of the neural network $g_{\boldsymbol{\phi_{\lambda}}}$. The variational distribution of $\boldsymbol{z}_{i,j}$ is also the Gaussian distribution and parameters of $q_{\boldsymbol{\phi_{z}}}(\boldsymbol{z}_{i,j}|\boldsymbol{x}_{i,j})$ are derived from the neural network $g_{\phi_{z}}$. For inference, only cell images $\boldsymbol{x}_{\neg 3,3}$ in the context panel are available. The most essential part of the variational distribution is $q_{\boldsymbol{\theta}}(\boldsymbol{z}_{3,3}^{d}|\boldsymbol{z}_{\neg 3,3}^{d},\boldsymbol{\lambda}^{row},\boldsymbol{\lambda}^{col})$ which follows the same calculation manner as the noise-free prediction with GPs. Denoting the Gaussian distribution of $\boldsymbol{z}_{3,3}^{d}$ as $\mathcal{N}\left(\mu_{\boldsymbol{\theta}}, \sigma_{\boldsymbol{\theta}}^{2}\right)$, where
\begin{equation}
\begin{split}
    \mu_{\boldsymbol{\theta}} &= \boldsymbol{R}_{\boldsymbol{\theta}}^{\top} \boldsymbol{C}_{\boldsymbol{\theta}}^{-1} \boldsymbol{t}, \quad \sigma_{\boldsymbol{\theta}}^{2} = d_{\boldsymbol{\theta}}-\boldsymbol{R}_{\boldsymbol{\theta}}^{\top} \boldsymbol{C}_{\boldsymbol{\theta}}^{-1} \boldsymbol{R}_{\boldsymbol{\theta}}, \\
    \boldsymbol{\lambda}_{1:9} &= flatten\left(\boldsymbol{\lambda}\right), \quad
    \boldsymbol{z}_{1:9}^{d} = flatten\left(\boldsymbol{z}^{d}\right) \\
    \boldsymbol{C}_{\boldsymbol{\theta}} &= \left[\begin{array}{ccc}
    \kappa_{\boldsymbol{\theta}}^{d}\left(\boldsymbol{\lambda}_{1}, \boldsymbol{\lambda}_{1}\right) & \cdots & \kappa_{\boldsymbol{\theta}}^{d}\left(\boldsymbol{\lambda}_{1}, \boldsymbol{\lambda}_{8}\right) \\
    \vdots & \ddots & \vdots \\
    \kappa_{\boldsymbol{\theta}}^{d}\left(\boldsymbol{\lambda}_{8}, \boldsymbol{\lambda}_{1}\right) & \cdots & \kappa_{\boldsymbol{\theta}}^{d}\left(\boldsymbol{\lambda}_{8}, \boldsymbol{\lambda}_{8}\right)
    \end{array}\right] \\
    \boldsymbol{R}_{\boldsymbol{\theta}} &= \left[\kappa_{\boldsymbol{\theta}}^{d}\left(\boldsymbol{\lambda}_{1}, \boldsymbol{\lambda}_{9}\right), ..., \kappa_{\boldsymbol{\theta}}^{d}\left(\boldsymbol{\lambda}_{8}, \boldsymbol{\lambda}_{9}\right) \right]^{\top} \\
    \boldsymbol{t} &= \left[z_{1}^{d}, ..., z_{8}^{d} \right]^{\top}, \quad d_{\boldsymbol{\theta}} = \kappa_{\boldsymbol{\theta}}^{d}\left(\boldsymbol{\lambda}_{9}, \boldsymbol{\lambda}_{9}\right)
\end{split}
\label{eq:inference}
\end{equation}
To simplify the above equation, we omit the dimension symbol $d$ in matrices. Eq.(\ref{eq:inference}) indicates that when inferring $z_{3,3}^{d}$, we need first to obtain $\boldsymbol{z}_{\neg3,3}^{d}$, $\boldsymbol{\lambda}^{row}$ and $\boldsymbol{\lambda}^{col}$ from cell images in the context panel. Considering the $\boldsymbol{\lambda}^{row}$ and $\boldsymbol{\lambda}^{col}$ as locations and $\boldsymbol{z}_{\neg3,3}^{d}$ as function values in the rule function, the distribution over $z_{3,3}^{d}$ can be eventually acquired.

\subsection{Parameter Learning}

In this paper, we optimize the ELBO $\mathcal{L}$ instead of the log-likelihood $
\log p_{\boldsymbol{\varphi}, \boldsymbol{\theta}}\left(\boldsymbol{x}\right)$ by amortized variational inference.
\begin{equation}
\begin{split}
    \mathcal{L} &= \mathbb{E}_{q_{\boldsymbol{\phi}, \boldsymbol{\theta}}\left(\boldsymbol{\eta}\mid\boldsymbol{x}_{\neg3,3}\right)}\left[\log \frac{p_{\boldsymbol{\varphi}, \boldsymbol{\theta}}\left(\boldsymbol{x},\boldsymbol{\eta}\right)}{q_{\boldsymbol{\phi}, \boldsymbol{\theta}}\left(\boldsymbol{\eta}\mid\boldsymbol{x}_{\neg3,3}\right)}\right] \\
    &= \mathbb{E}_{q_{\boldsymbol{\phi}, \boldsymbol{\theta}}\left(\boldsymbol{z}, \boldsymbol{\lambda} \mid \boldsymbol{x}_{\neg3,3}\right)}\biggr[\log p_{\boldsymbol{\varphi}}\left(\boldsymbol{x} \mid \boldsymbol{z}\right)\biggr] \\
    &\quad - \beta D_{KL}\left(q_{\boldsymbol{\phi}}\left(\boldsymbol{\lambda} \mid \boldsymbol{x}_{\neg3,3}\right) \| p(\boldsymbol{\lambda})\right) \\
    &\quad - \gamma \mathbb{E}_{q_{\boldsymbol{\phi}}\left(\boldsymbol{\lambda} \mid \boldsymbol{x}_{\neg3,3}\right)}\biggr[\sum_{d=1}^{D}D_{KL}\Big(q_{\boldsymbol{\theta}}\left(z_{3,3}^{d} \mid \boldsymbol{z}_{\neg3,3}^{d}, \boldsymbol{\lambda}\right) \\
    & \quad \quad q_{\boldsymbol{\phi}}\left(\boldsymbol{z}_{\neg3,3}^{d} \mid \boldsymbol{x}_{\neg3,3}\right) \| p_{\boldsymbol{\theta}}\left(\boldsymbol{z}^{d} \mid \boldsymbol{\lambda}\right)\Big)\biggr]
\end{split}
\label{eq:elbo}
\end{equation}
where $\beta$ and $\gamma$ are hyperparameters employed to control the penalty in KL divergence to impose the dimensional independence on latent variables \cite{higgins2016beta}. To estimate the expectation of the first and third terms in Eq.(\ref{eq:elbo}), we sample $\boldsymbol{\lambda}$ and $\boldsymbol{z}$ from the variational posterior with reparameterization trick \cite{kingma2013auto}. Then the gradient-based backpropagation algorithm is executed to adjust the parameters of involved neural networks. However, when calculating KL divergence in the third term, the explicit form of distribution $q_{\boldsymbol{\theta}}(z_{3,3}^{d}|\boldsymbol{z}_{\neg3,3}^{d}, \boldsymbol{\lambda})q_{\boldsymbol{\phi}}(\boldsymbol{z}_{\neg3,3}^{d}|\boldsymbol{x}_{\neg3,3})$ is hard to derive. Alternatively, as the marginal distributions of $p_{\boldsymbol{\theta}}(\boldsymbol{z}^{d}|\boldsymbol{\lambda})$ are computationally tractable, we propose a marginal matching method to approximate the non-computable KL divergence. To match marginal distributions of the posterior and prior, we substitute the prior with the product of its marginal distributions $p_{\boldsymbol{\theta}}(z_{3,3}^{d}|\boldsymbol{z}_{\neg3,3}^{d}, \boldsymbol{\lambda})p_{\boldsymbol{\theta}}(\boldsymbol{z}_{\neg3,3}^{d}|\boldsymbol{\lambda})$. By means of this approximation method, the third term can be transformed into a summary over the KL divergence between marginal distributions
\begin{equation}
\begin{split}
    \sum_{(i,j) \in \Omega}D_{KL}\left(q_{\boldsymbol{\phi}}\left(z_{i,j}^{d} \mid \boldsymbol{x}_{i,j} \right) \| p_{\boldsymbol{\theta}}\left(z_{i,j}^{d} \mid \boldsymbol{\lambda} \right )\right)
\end{split}
\end{equation}
By maximizing the approximated ELBO, our model can be trained end to end. The detailed derivation process is described in the supplementary material.
    
\section{Experiments}

\subsection{RPM-like Datasets}

We create new RPM-like datasets that remove selection panels and retain complete 9-cell panels. Visual concepts of the panel object in a dataset are divided into a changeable and unchangeable concept set. Concepts in the changeable concept set could change with rules but those in the unchangeable concept cannot. All concepts change in continuous spaces, and rules can be exerted in both the horizontal and vertical directions. Datasets are generated according to the following rules:
\begin{itemize}
	\item \textbf{progress}: one concept changes with isometric steps horizontally or vertically (2nd ground truth in Figure \ref{fig:exp_panel_generate}(a));
	\item \textbf{biprogress}: one concept changes with isometric steps in \textit{both} directions (1st ground truth in Figure \ref{fig:exp_panel_generate}(a));
	\item \textbf{multiprogress}: \textit{multiple concepts} change with isometric steps horizontally or vertically (1st and 2nd ground truths in Figure \ref{fig:exp_panel_generate}(b)).
\end{itemize}
To create a panel, we first randomly choose a rule and a subset of the changeable concept set. As Figure \ref{fig:exp_panel_generate} shows, the values of the chosen concept across rows are stochastically determined under constraints from the selected rule. The remaining concepts are kept fixed in a row (we consider only horizontal rules now). Finally, the panel will be transposed with the 50\% probability to produce vertical rules.

In the following experiments, our model will be evaluated in two instances of the RPM-like datasets:
\begin{itemize}
    \item \textbf{Polygon}: each cell has a centered polygon (one dataset contains only one type of polygon, and we choose the triangle here) with the changeable concepts $size$ and $grayscale$ as well as the unchangeable $rotation$. Panels can follow the rules $progress$ and $biprogress$.
    \item \textbf{Circle}: there is a centered hollow circle in each panel cell with a black sector varied in the start position and size.  We call the start position and size as the changeable concepts $position$ and $size$, respectively, and there is no unchangeable concept. In the Circle dataset, we introduce the rules $progress$ and $multiprogress$.
\end{itemize}
Results about more instances of the RPM-like dataset and the selective task accuracy compared with RPM solvers are exhibited in the supplementary material.

Each instance of the dataset contains a disjoint training set, validation set, and test set. For further performance evaluation, we provide various-sized (from 50 to 50000) training sets and one fixed 10000-sample test set.

\subsection{Quantitative Results}

\begin{figure}[t]
\centering
\includegraphics[width=0.7\columnwidth]{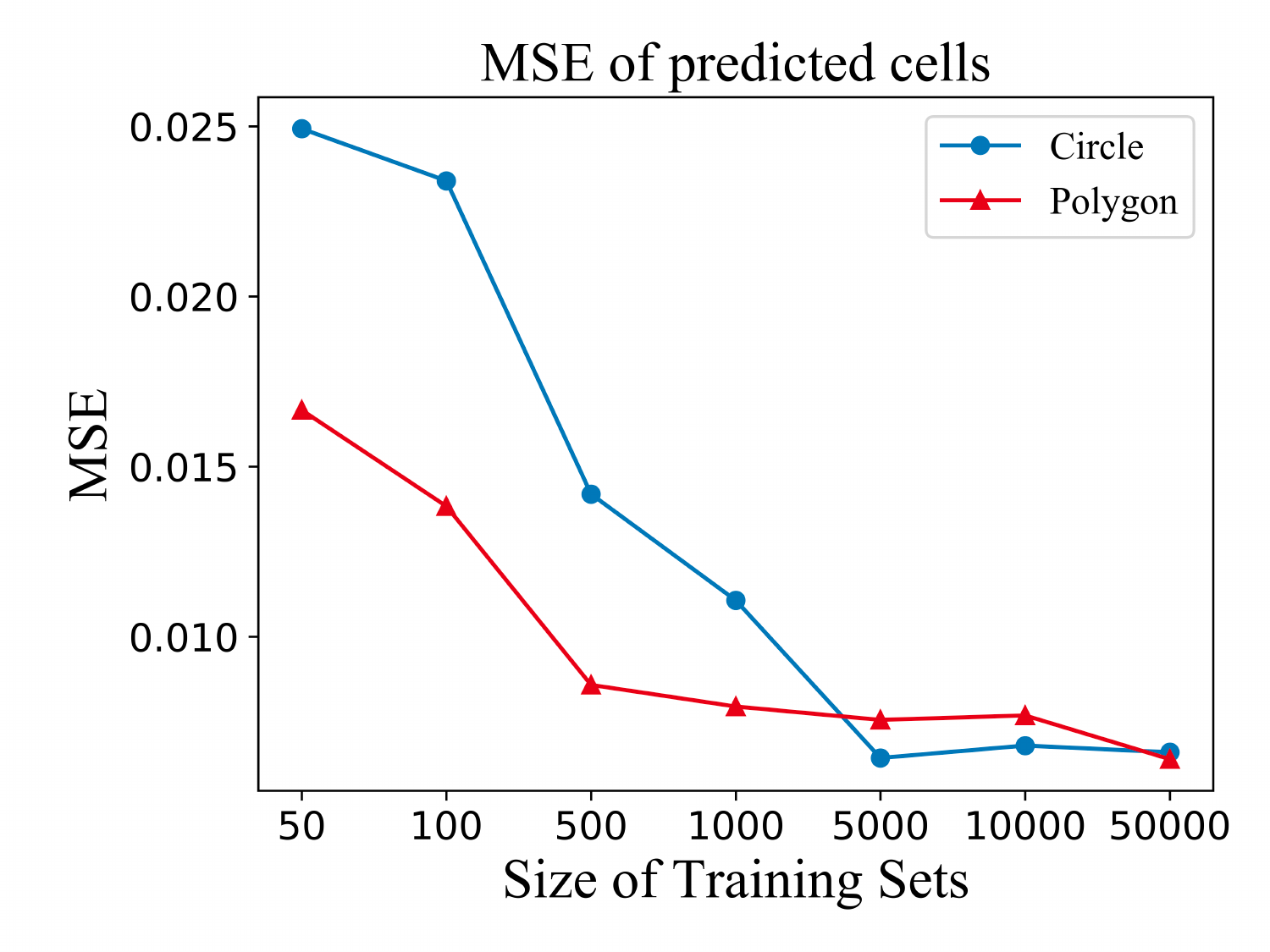}
\caption{MSEs of our model on different-sized training sets.}
\label{fig:mse}
\end{figure}

\begin{table}[t]
\centering
\begin{tabular}{c|c|c}
\hline
 & Polygon & Circle \\ \hline
VAE-GAN & 5.16e-2 & 1.08e-1 \\ \hline
Proposed & \textbf{7.55e-3 $\pm$ 3.9e-5} & \textbf{6.43e-3 $\pm$ 1.8e-5} \\ \hline
\end{tabular}
\caption{MSEs of two models. Predictions of VAE-GAN are determinate, thus the repeated test is not performed.}
\label{tab:mse}
\end{table}

\textbf{Evaluation Metrics.} To assess the painting performance, we use the \textit{Mean Squared Error} (MSE) of predicted cell images. As only predicting right-bottom cells may neglect the relations between panel cells, we thus consider the varying-position MSE score. In a single prediction procedure, the tested model should predict a target cell given the rest 8 panel cells. The model will predict all 9 cells in turn, and calculate the average MSE between predictions and ground truths. This evaluation method estimates both the reconstruction error of predictions and the generalization ability to predict cells in novel positions.

\textbf{Data Efficiency.} By increasing the training set size from 50 to 50000, results in Figure \ref{fig:mse} display satisfying results of our model on the Polygon training set with only 500 samples, and Circle training set with only 5000 samples. This phenomenon suggests that our model is data-efficient, which is beneficial from the interpretable GP rule reasoning. Therefore, in the following experiment, the training set size is set to 5000 except for additional statements.

\textbf{Model Comparison.} Since there exist very few previous works in the RPM painting problem, we choose to compare with a VAE-GAN inpainting model (referred to as VAE-GAN) \cite{hua2019modeling}, which is applied in the missing region inpainting of images \cite{yu2018generative}. MSE scores on both two datasets are available in Table \ref{tab:mse} where our model outperforms VAE-GAN trained with 50000 samples, indicating that our model's painting ability can be automatically generalized from the right-bottom to other positions. In contrast, relatively low MSE scores of VAE-GAN implies its limited generalization ability for cell prediction. We provide the visualized prediction results of VAE-GAN in the supplementary material.

\subsection{Cell Prediction}

\begin{figure}[t]
\centering
\includegraphics[width=0.9\columnwidth]{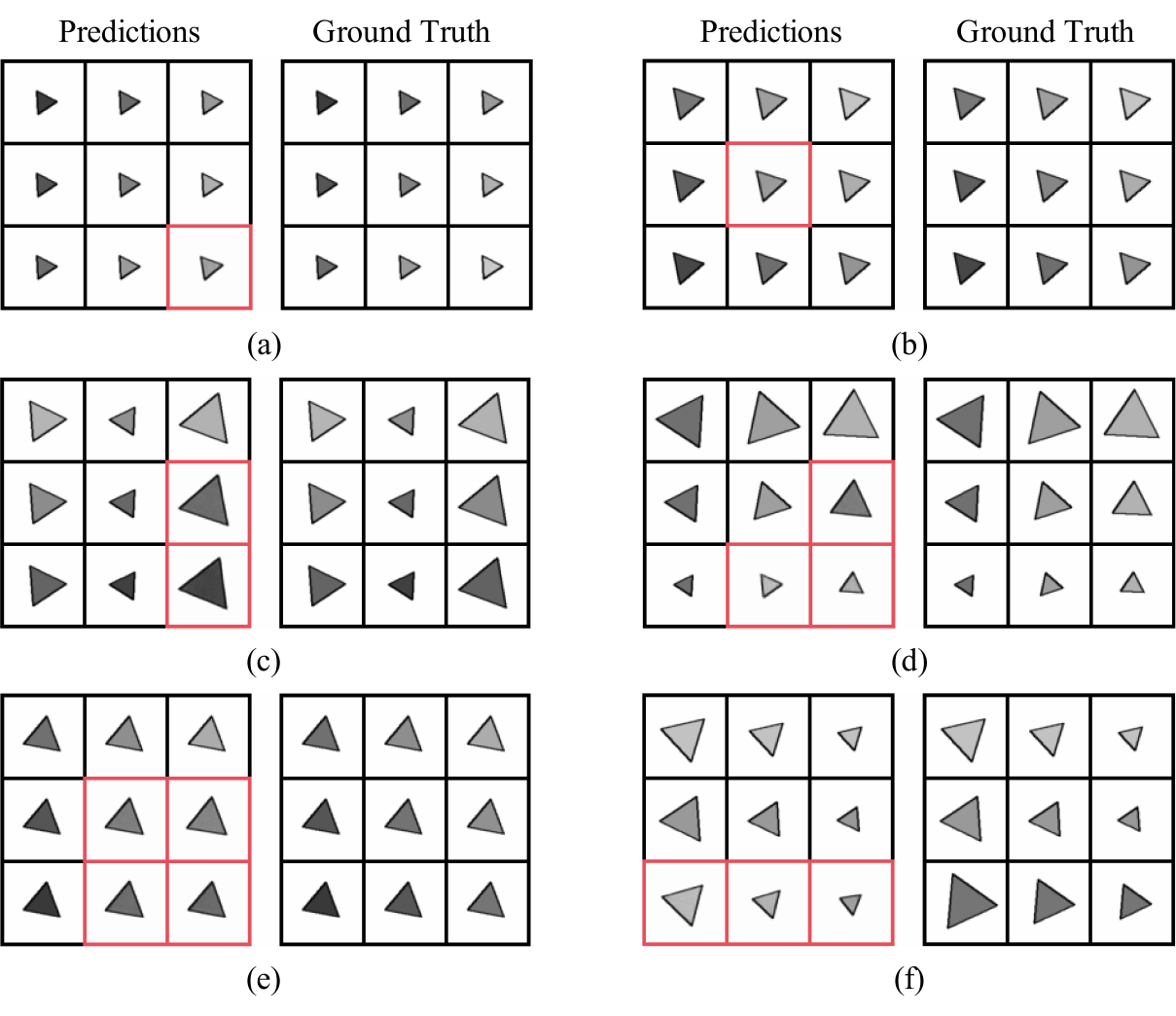}
\caption{Prediction results on the Polygon dataset. Cell images bounded by red boxes are simultaneously drawn (predicted) by our model while other cells contain ground truth images.}
\label{fig:exp_cell_prediction}
\end{figure}

For cell prediction, the varying-position and multi-cell prediction tasks emphasize the understanding of concept-changing rules. Figure \ref{fig:exp_cell_prediction} visualizes predictions of our model trained on the Polygon training set, which validates that our model can draw accurate predictions, and more importantly, understand concept-changing rules. Figure \ref{fig:exp_cell_prediction}(a) and \ref{fig:exp_cell_prediction}(b) substantially evidence the varying-position prediction ability of our model, that is, it is able to predict all cells in panels by training to predict only right-bottom cells. Furthermore, experiments about the multi-cell prediction in Figure \ref{fig:exp_cell_prediction}(c) to \ref{fig:exp_cell_prediction}(f) show that our model successfully captures the underlying concept-changing rules via GP priors. Predictions in Figure \ref{fig:exp_cell_prediction}(c) are closed to ground truth images. However, in Figure \ref{fig:exp_cell_prediction}(e), predictions deviate obviously in the concept $grayscale$. By further concealing a whole row from a panel as in Figure \ref{fig:exp_cell_prediction}(f), our model derives the horizontal rule $progress$ from remaining rows but the generated cell images have distinct sizes and colors compared to the ground truths. We deduce that there is enough context information for the concept-changing rule inference but inadequate information for the missing row reconstruction (because the whole row is removed). These results suggest that our model can execute robust and interpretable cell prediction.

\subsection{Panel Sampling and Interpolation}

Due to the explicit generative process that transforms cell latent variables $\boldsymbol{\lambda}$ to images $\boldsymbol{x}$, it is possible to construct novel panels. The novel panels will be sampled via the learned latent GP priors. Due to learnable parameters in the kernels, rule function spaces are well-adapted to the dataset. In Figure \ref{fig:exp_panel_generate}(a), generated panels are diverse: the vertical rule $progress$ (3rd sample), horizontal rule $progress$ (1st sample), and the rule $biprogress$ (4th sample); the changing concept $size$ (3rd sample) as well as $grayscale$ (4th sample). The concept $rotation$ of generated panels is not changed progressively in a row or column, which agrees to our definition about the unchangeable set. The generated panels in Figure \ref{fig:exp_panel_generate}(b) also satisfy the predefined creation principles about the Circle dataset.

For visual concepts, the latent GP priors offer a methodology for function value interpolation. Figure \ref{fig:exp_panel_generate} suggest that a larger panel can be generated by interpolation between cell latent variables. Owing to the smoothness of continuous functions, interpolated cell latent variables will be decoded to continuous-varying images. Three subfigures in the right of Figure \ref{fig:exp_panel_generate}(a) display interpolation results on the Polygon dataset. By interpolating between corner cells, progressive changes in the concepts $grayscale$ (all three samples) and $rotation$ (1st and 3rd samples) are achieved. Values of the concepts $position$ (1st and 2nd samples) and $size$ (2nd and 3rd samples) change smoothly as well when interpolating on the Circle dataset.

\subsection{Interpretable Visual Concepts}

\begin{figure}[t]
\centering
\includegraphics[width=0.99\columnwidth]{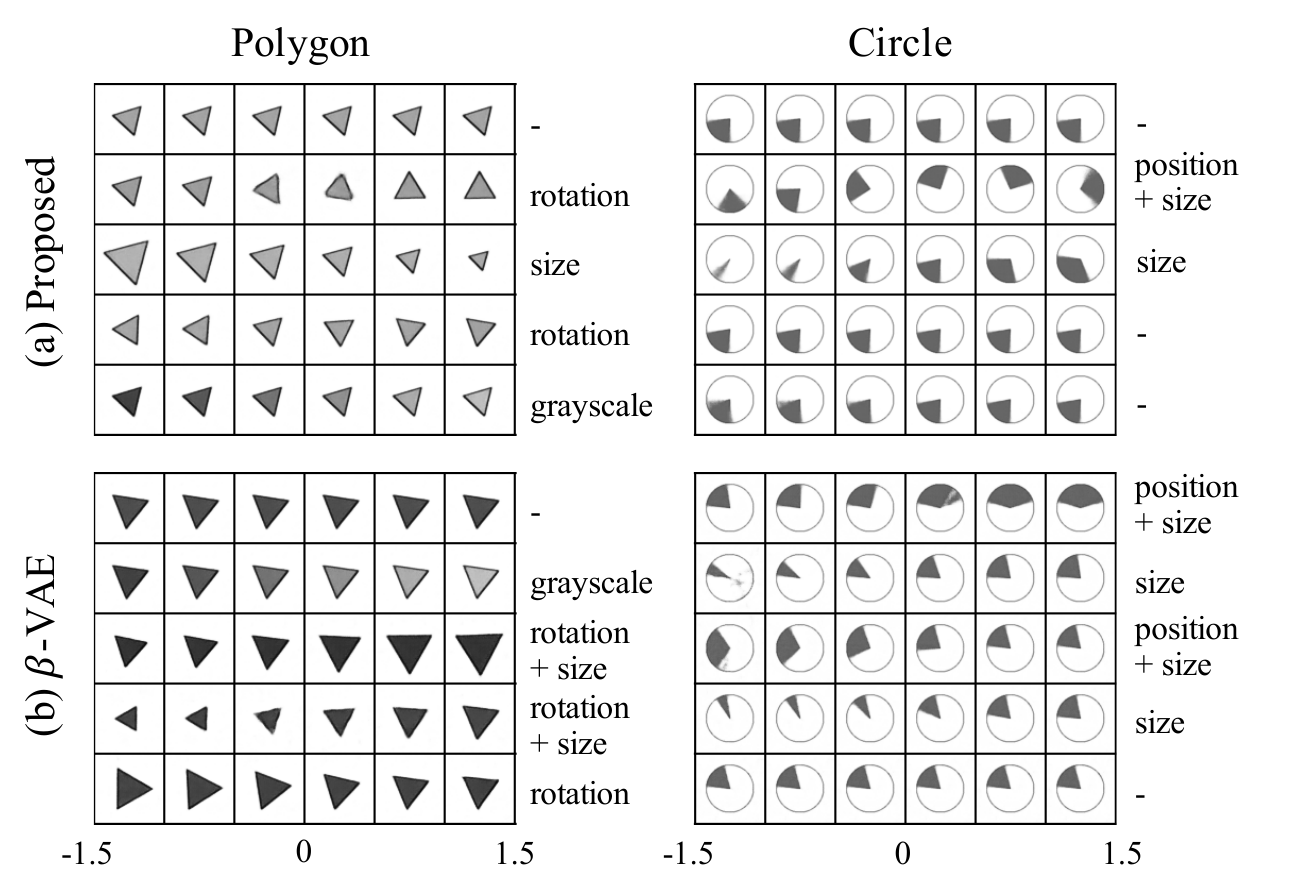}
\caption{Traverse the dimensions of a cell image latent variable ranged from -1.5 to 1.5 on two datasets. On the right side of images, related visual concepts of each dimension are listed.}
\label{fig:disentangle}
\end{figure}

\begin{table}[t]
\centering
\begin{tabular}{|c|c|c|}
\hline
\multicolumn{3}{|c|}{Factor VAE Score} \\ 
\cline{1-3} & Polygon & Circle \\ \hline
$\beta$-VAE & \textbf{1.00 $\pm$ 0.0} & 7.59e-1 $\pm$ 1.8e-2 \\ \hline
Proposed & \textbf{1.00 $\pm$ 0.0} & \textbf{1.00 $\pm$ 0.0} \\ \hline \hline
\multicolumn{3}{|c|}{SAP Score} \\
\cline{1-3} & Polygon & Circle \\ \hline
$\beta$-VAE & 4.40e-1 $\pm$ 6.5e-3 & 9.41e-2 $\pm$ 3.0e-3 \\ \hline
Proposed & \textbf{7.80e-1 $\pm$ 3.8e-3} & \textbf{7.93e-1 $\pm$ 4.0e-3} \\
\hline
\end{tabular}
\caption{Factor VAE Scores and SAP Scores of the $\beta$-VAE and our model.}
\label{tab:disentangle}
\end{table}

The independence and interpretability of cell image latent variables are essential since the GP rule reasoning is respectively exerted in each visual concept. Disentangled models are considered as a way to impose independence and meanings to dimensions. Because hyperparameters $\beta$ and $\gamma$ within Eq.(\ref{eq:elbo}) are similar to those in $\beta$-VAE loss, we compare our model with $\beta$-VAE  \cite{higgins2016beta}  to investigate the effect of the latent GP priors. $\beta$-VAE is trained with 50000 samples to avoid insufficient training.

\textbf{Evaluation Metrics.} \textit{Factor VAE Score} \cite{kim2018disentangling} and \textit{SAP Score} \cite{kumar2017variational} are exploited as evaluation metrics. Factor VAE Score focuses on the modularity among dimensions by using a majority-vote classifier to map dimensions to visual concepts. When computing SAP Score, a correlation matrix of all dimensions and concepts is computed to acquire the difference between top-2 scores, which stresses the code compactness of representations. In short, to get a high Factor VAE Score, we should ensure one dimension to encode at most one concept. If a superior SAP Score is expected, the model has better code a concept with only one dimension. Since two scores inspect distinct aspects of the representation independence and interpretability, we integrate them to comprehensively assess the performance of our model.

Table \ref{tab:disentangle} indicates that our model achieves higher scores than $\beta$-VAE on the Polygon and Circle datasets. Figure \ref{fig:disentangle} gives more details through the dimension-wise latent space traverse. When evaluating our model on the Polygon dataset, nearly all dimensions encode one or less concept. Though Factor VAE Score of $\beta$-VAE on the Polygon dataset is competitive, visualized results suggest that the 3rd and 4th dimensions encode two concepts together. We consider that perturbations over the 5th dimension impose more deviation on the concept $rotation$ compared with the 3rd and 4th dimensions, which supplies clues for the classifier to find the correct mapping. SAP Scores of two models are lower than Factor VAE Scores because both models fail to code one concept with merely one dimension. On the Circle dataset, our model encodes the concept $size$ into two dimensions (2nd and 3rd). However, $\beta$-VAE applies four dimensions (1st to 4th) for the concept $size$ and two dimensions (1st and 3rd) for $position$. It is evidenced by the above experiments that the GP rule reasoning allows our model to learn more reliable representations whose dimensions are independent and correspond to visual concepts (or have no meaning). The latent GP priors reveal the concept-changing rules in panels and bring more inductive biases for better disentanglement.

\section{Conclusion}

In this paper, we propose an interpretable deep latent variable model to solve the RPM painting problem. Our model is capable of learning visual concepts from datasets, GP rule reasoning in latent variable spaces, and painting missing cell images in RPM panels. For the GP rule reasoning, we define underlying concept-changing rules as 2D functions whose priors are the latent GPs. Independent and interpretable latent variables facilitate the dimension-wise abstract reasoning, while inference with entire panels in turn helps the learning of visual concepts. Due to the latent GP priors, the proposed model is not exclusively limited in predicting cell contents, it is able to naturally generate RPM panels. 

\section{Acknowledgments}

This research was supported in part by STCSM Projects (20511100400, 18511103104), Shanghai Municipal Science and Technology Major Projects (2017SHZDZX01, 2018SHZDZX01), Shanghai Research and Innovation Functional Program (17DZ2260900), and the Program for Professor of Special Appointment (Eastern Scholar) at Shanghai Institutions of Higher Learning.

\bibliography{content.bib}

\clearpage
\newpage
\onecolumn

\section{Evidence Lower Bound}

\begin{equation}
\begin{split}
    \log p_{\boldsymbol{\varphi},\boldsymbol{\theta}}\left(\boldsymbol{x}\right) &\geq \log p_{\boldsymbol{\varphi},\boldsymbol{\theta}}\left(\boldsymbol{x}\right) - D_{KL}\left(q_{\boldsymbol{\phi},\boldsymbol{\theta}}\left(\boldsymbol{l}|\boldsymbol{x}_{\neg3,3}\right)\|p_{\boldsymbol{\varphi},\boldsymbol{\theta}}\left(\boldsymbol{l}|\boldsymbol{x}\right)\right) \nonumber \\
    &=E_{q_{\boldsymbol{\phi},\boldsymbol{\theta}}\left(\boldsymbol{l}|\boldsymbol{x}_{\neg3,3}\right)}\left[\log\frac{p_{\boldsymbol{\varphi},\boldsymbol{\theta}}\left(\boldsymbol{l},\boldsymbol{x}\right)}{q_{\boldsymbol{\phi},\boldsymbol{\theta}}\left(\boldsymbol{l}|\boldsymbol{x}_{\neg3,3}\right)}\right] \\
    &=E_{q_{\boldsymbol{\phi},\boldsymbol{\theta}}\left(\boldsymbol{l}|\boldsymbol{x}_{\neg3,3}\right)}\left[\log\frac{p\left(\boldsymbol{\lambda}\right)p_{\boldsymbol{\theta}}\left(\boldsymbol{z}|\boldsymbol{\lambda}\right)p_{\boldsymbol{\varphi}}\left(\boldsymbol{x}|\boldsymbol{z}\right)}{\prod_{d=1}^{D}q_{\boldsymbol{\theta}}\left(\boldsymbol{z}_{3,3}^{d}|\boldsymbol{z}_{\neg3,3}^{d},\boldsymbol{\lambda}\right)q_{\boldsymbol{\phi}}\left(\boldsymbol{\lambda}|\boldsymbol{x}_{\neg3,3}\right)q_{\boldsymbol{\phi}}\left(\boldsymbol{z}_{\neg3,3}|\boldsymbol{x}_{\neg3,3}\right)}\right] \\
    &=E_{q_{\boldsymbol{\phi},\boldsymbol{\theta}}\left(\boldsymbol{l}|\boldsymbol{x}_{\neg3,3}\right)}\left[\log p_{\boldsymbol{\varphi}}\left(\boldsymbol{x}|\boldsymbol{z}\right)\right]
    + E_{q_{\boldsymbol{\phi},\boldsymbol{\theta}}\left(\boldsymbol{l}|\boldsymbol{x}_{\neg3,3}\right)}\left[\log\frac{p\left(\boldsymbol{\lambda}\right)}{q_{\boldsymbol{\phi}}\left(\boldsymbol{\lambda}|\boldsymbol{x}_{\neg3,3}\right)}\right] \\
    & \quad\quad + E_{q_{\boldsymbol{\phi},\boldsymbol{\theta}}\left(\boldsymbol{l}|\boldsymbol{x}_{\neg3,3}\right)}\left[\log\frac{\prod_{d=1}^{D}p_{\boldsymbol{\theta}}\left(\boldsymbol{z}^{d}|\boldsymbol{\lambda}\right)}{\prod_{d=1}^{D}q_{\boldsymbol{\theta}}\left(\boldsymbol{z}_{3,3}^{d}|\boldsymbol{z}_{\neg3,3}^{d},\boldsymbol{\lambda}\right)\prod_{d=1}^{D}q_{\boldsymbol{\phi}}\left(\boldsymbol{z}_{\neg3,3}^{d}|\boldsymbol{x}_{\neg3,3}\right)}\right] \\
    &=E_{q_{\boldsymbol{\phi},\boldsymbol{\theta}}\left(\boldsymbol{l}|\boldsymbol{x}_{\neg3,3}\right)}\left[\log p_{\boldsymbol{\varphi}}\left(\boldsymbol{x}|\boldsymbol{z}\right)\right]
    + E_{q_{\boldsymbol{\phi}}\left(\boldsymbol{\lambda}|\boldsymbol{x}_{\neg3,3}\right)}\left[E_{q_{\boldsymbol{\phi},\boldsymbol{\theta}}\left(\boldsymbol{z}|\boldsymbol{\lambda},\boldsymbol{x}_{\neg3,3}\right)}\left[\log\frac{p\left(\boldsymbol{\lambda}\right)}{q_{\boldsymbol{\phi}}\left(\boldsymbol{\lambda}|\boldsymbol{x}_{\neg3,3}\right)}\right]\right]\\
    & \quad\quad + E_{q_{\boldsymbol{\phi}}\left(\boldsymbol{\lambda}|\boldsymbol{x}_{\neg3,3}\right)}\left[E_{q_{\boldsymbol{\phi},\boldsymbol{\theta}}\left(\boldsymbol{z}|\boldsymbol{\lambda},\boldsymbol{x}_{\neg3,3}\right)}\left[\sum_{d=1}^{D}\log\frac{p_{\boldsymbol{\theta}}\left(\boldsymbol{z}^{d}|\boldsymbol{\lambda}\right)}{q_{\boldsymbol{\theta}}\left(\boldsymbol{z}_{3,3}^{d}|\boldsymbol{z}_{\neg3,3}^{d},\boldsymbol{\lambda}\right)q_{\boldsymbol{\phi}}\left(\boldsymbol{z}_{\neg3,3}^{d}|\boldsymbol{x}_{\neg3,3}\right)}\right]\right] \\
    &=E_{q_{\boldsymbol{\phi},\boldsymbol{\theta}}\left(\boldsymbol{l}|\boldsymbol{x}_{\neg3,3}\right)}\left[\log p_{\boldsymbol{\varphi}}\left(\boldsymbol{x}|\boldsymbol{z}\right)\right]
    - D_{KL}\left(q_{\boldsymbol{\phi}}\left(\boldsymbol{\lambda}|\boldsymbol{x}_{\neg3,3}\right)\|p\left(\boldsymbol{\lambda}\right)\right) \\
    & \quad\quad - E_{q_{\boldsymbol{\phi}}\left(\boldsymbol{\lambda}|\boldsymbol{x}_{\neg3,3}\right)}\left[\sum_{d=1}^{D}D_{KL}\left(q_{\boldsymbol{\theta}}\left(\boldsymbol{z}_{3,3}^{d}|\boldsymbol{z}_{\neg3,3}^{d},\boldsymbol{\lambda}\right)q_{\boldsymbol{\phi}}\left(\boldsymbol{z}_{\neg3,3}^{d}|\boldsymbol{x}_{\neg3,3}\right)\|p_{\boldsymbol{\theta}}\left(\boldsymbol{z}^{d}|\boldsymbol{\lambda}\right)\right)\right]
\end{split}
\end{equation}

\section{Marginal Matching Approximation}

\begin{equation}
\begin{split}
    D_{KL}&\left(q_{\boldsymbol{\theta}}\left(\boldsymbol{z}_{3,3}^{d}|\boldsymbol{z}_{\neg3,3}^{d},\boldsymbol{\lambda}\right)q_{\boldsymbol{\phi}}\left(\boldsymbol{z}_{\neg3,3}^{d}|\boldsymbol{x}_{\neg3,3}\right)\|p_{\boldsymbol{\theta}}\left(\boldsymbol{z}^{d}|\boldsymbol{\lambda}\right)\right) \nonumber\\
    &= -E_{q_{\boldsymbol{\theta}}\left(\boldsymbol{z}_{3,3}^{d}|\boldsymbol{z}_{\neg3,3}^{d},\boldsymbol{\lambda}\right)q_{\boldsymbol{\phi}}\left(\boldsymbol{z}_{\neg3,3}^{d}|\boldsymbol{x}_{\neg3,3}\right)}\left[\log\frac{p_{\boldsymbol{\theta}}\left(\boldsymbol{z}^{d}|\boldsymbol{\lambda}\right)}{q_{\boldsymbol{\theta}}\left(\boldsymbol{z}_{3,3}^{d}|\boldsymbol{z}_{\neg3,3}^{d},\boldsymbol{\lambda}\right)q_{\boldsymbol{\phi}}\left(\boldsymbol{z}_{\neg3,3}^{d}|\boldsymbol{x}_{\neg3,3}\right)}\right] \\
    &= -E_{q_{\boldsymbol{\theta}}\left(\boldsymbol{z}_{3,3}^{d}|\boldsymbol{z}_{\neg3,3}^{d},\boldsymbol{\lambda}\right)q_{\boldsymbol{\phi}}\left(\boldsymbol{z}_{\neg3,3}^{d}|\boldsymbol{x}_{\neg3,3}\right)}\left[\log\frac{p_{\boldsymbol{\theta}}\left(\boldsymbol{z}_{3,3}^{d}|\boldsymbol{z}_{\neg3,3}^{d},\boldsymbol{\lambda}\right)p_{\boldsymbol{\theta}}\left(\boldsymbol{z}_{\neg3,3}^{d}|\boldsymbol{\lambda}\right)}{q_{\boldsymbol{\theta}}\left(\boldsymbol{z}_{3,3}^{d}|\boldsymbol{z}_{\neg3,3}^{d},\boldsymbol{\lambda}\right)q_{\boldsymbol{\phi}}\left(\boldsymbol{z}_{\neg3,3}^{d}|\boldsymbol{x}_{\neg3,3}\right)}\right] \\
    &= -E_{q_{\boldsymbol{\phi}}\left(\boldsymbol{z}_{\neg3,3}^{d}|\boldsymbol{x}_{\neg3,3}\right)}\left[E_{q_{\boldsymbol{\theta}}\left(\boldsymbol{z}_{3,3}^{d}|\boldsymbol{z}_{\neg3,3}^{d},\boldsymbol{\lambda}\right)}\left[\log\frac{p_{\boldsymbol{\theta}}\left(\boldsymbol{z}_{\neg3,3}^{d}|\boldsymbol{\lambda}\right)}{q_{\boldsymbol{\phi}}\left(\boldsymbol{z}_{\neg3,3}^{d}|\boldsymbol{x}_{\neg3,3}\right)}\right]\right] \\
    &\approx -E_{q_{\boldsymbol{\phi}}\left(\boldsymbol{z}_{\neg3,3}^{d}|\boldsymbol{x}_{\neg3,3}\right)}\left[\log\frac{\prod_{(i,j)\in\Omega}p_{\boldsymbol{\theta}}\left(\boldsymbol{z}_{i,j}^{d}|\boldsymbol{\lambda}\right)}{\prod_{(i,j)\in\Omega}q_{\boldsymbol{\phi}}\left(\boldsymbol{z}_{i,j}^{d}|\boldsymbol{x}_{i,j}\right)}\right] \\
    &= -E_{q_{\boldsymbol{\phi}}\left(\boldsymbol{z}_{\neg3,3}^{d}|\boldsymbol{x}_{\neg3,3}\right)}\left[\sum_{(i,j)\in\Omega}\log\frac{p_{\boldsymbol{\theta}}\left(\boldsymbol{z}_{i,j}^{d}|\boldsymbol{\lambda}\right)}{q_{\boldsymbol{\phi}}\left(\boldsymbol{z}_{i,j}^{d}|\boldsymbol{x}_{i,j}\right)}\right] \\
    &= \sum_{(i,j)\in\Omega}D_{KL}\left(q_{\boldsymbol{\phi}}\left(\boldsymbol{z}_{i,j}^{d}|\boldsymbol{x}_{i,j}\right)\|p_{\boldsymbol{\theta}}\left(\boldsymbol{z}_{i,j}^{d}|\boldsymbol{\lambda}\right)\right)
\end{split}
\end{equation}
To calculate the marginal matching approximated Kullback-Leibler divergence, we sample $\boldsymbol{\lambda}$ to estimate $p_{\boldsymbol{\theta}}(\boldsymbol{z}_{i,j}^{d}|\boldsymbol{\lambda})$ with the reparameterization trick for gradient backpropagation. In the second step, the posterior and prior execute the identical Gaussian process rule reasoning, hence we have $p_{\boldsymbol{\theta}}(\boldsymbol{z}_{3,3}^{d}|\boldsymbol{z}_{\neg3,3}^{d},\boldsymbol{\lambda}) = q_{\boldsymbol{\theta}}(\boldsymbol{z}_{3,3}^{d}|\boldsymbol{z}_{\neg3,3}^{d},\boldsymbol{\lambda})$ for the same $\boldsymbol{\lambda}$, $\boldsymbol{z}_{\neg3,3}^{d}$ and parameter $\boldsymbol{\theta}$.

\section{Dataset}

\begin{figure*}[t]
  \centering
  \includegraphics[width=0.65\textwidth]{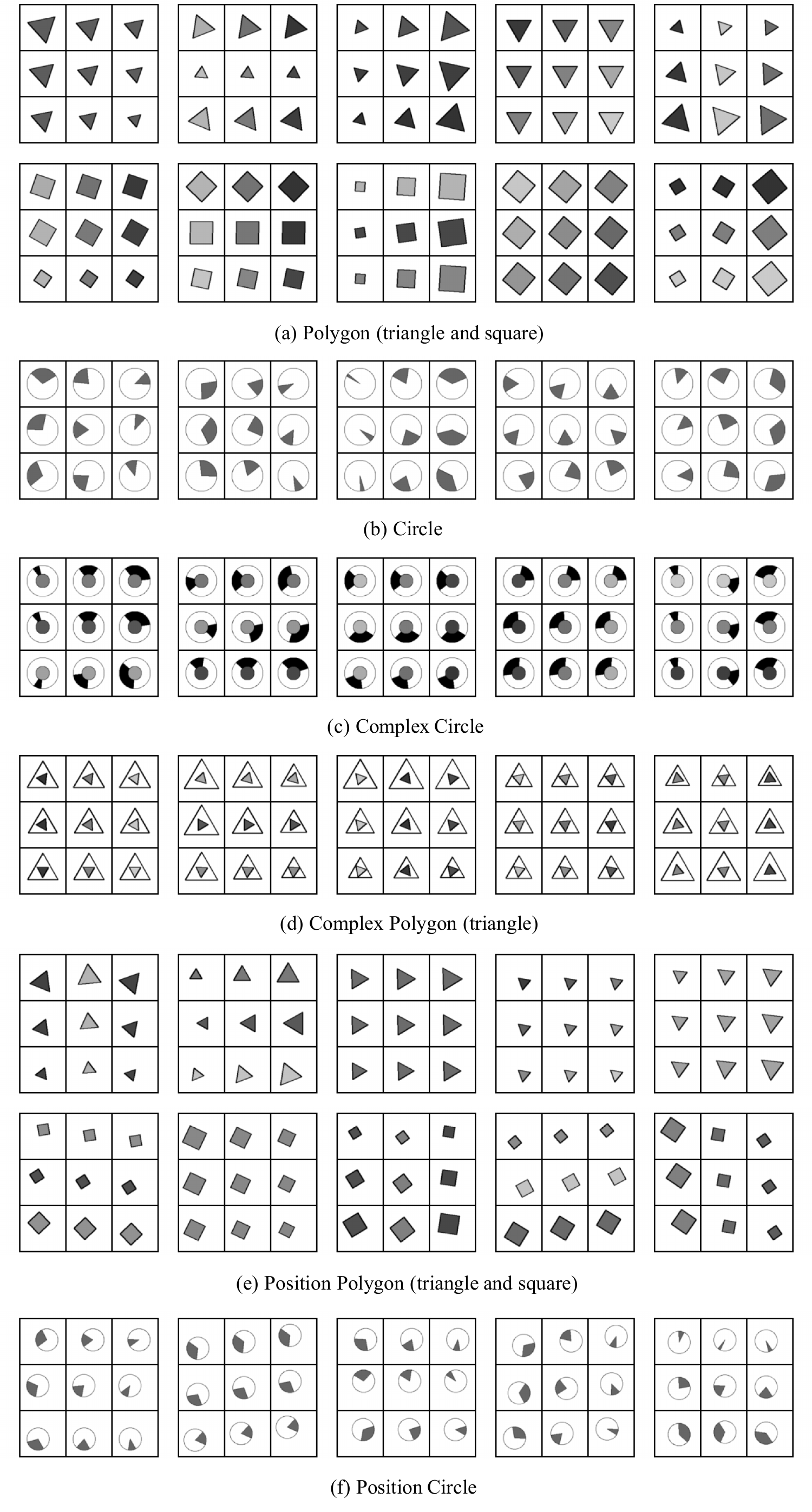}
  \caption{Samples from the RPM-like continuous-changing datasets.}
  \label{fig:sup_dataset}
\end{figure*}

\begin{table*}[t]
\centering
\begin{tabular}{|l|l|}
\hline
\multicolumn{1}{|c|}{\textbf{Polygon}} & \multicolumn{1}{c|}{\textbf{Circle}} \\ \hline
\multicolumn{2}{|c|}{\textbf{Rule Set}} \\ \hline
\textit{progress}, \textit{biprogress} & \textit{progress}, \textit{multiprogress} \\ \hline
\multicolumn{2}{|c|}{\textbf{Changeable Concept Set}} \\ \hline
\textit{grayscale}: the grayscale of centered polygons & \textit{position}: the start position of black sectors \\
\textit{size}: the size of centered polygons &
\textit{size}: the arc of black sectors \\ \hline
\multicolumn{2}{|c|}{\textbf{Unchangeable Concept Set}} \\
\hline
\textit{rotation}: the rotation of centered polygons & $\varnothing$ \\ \hline \hline

\multicolumn{1}{|c|}{\textbf{Complex Polygon}} & \multicolumn{1}{c|}{\textbf{Complex Circle}} \\ \hline
\multicolumn{2}{|c|}{\textbf{Rule Set}} \\ \hline
\textit{progress} & \textit{progress} \\ \hline
\multicolumn{2}{|c|}{\textbf{Changeable Concept Set}} \\ \hline
\textit{grayscale}: the grayscale of inner polygons & \textit{position}: the start position of black sectors \\
\textit{size}: the size of outer polygons & \textit{size}: the arc of black sectors \\ 
& \textit{grayscale}: the grayscale of inner circles \\ \hline
\multicolumn{2}{|c|}{\textbf{Unchangeable Concept Set}} \\
\hline
\textit{rotation}: the rotation of inner polygons & $\varnothing$ \\ \hline \hline

\multicolumn{1}{|c|}{\textbf{Position Polygon}} & \multicolumn{1}{c|}{\textbf{Position Circle}} \\ \hline
\multicolumn{2}{|c|}{\textbf{Rule Set}} \\ \hline
\textit{progress}, \textit{biprogress} & \textit{progress}, \textit{multiprogress} \\ \hline
\multicolumn{2}{|c|}{\textbf{Changeable Concept Set}} \\ \hline
\textit{grayscale}: the grayscale of inner polygons & \textit{position}: the start position of black sectors \\
\textit{size}: the size of outer polygons & \textit{size}: the arc of black sectors \\ 
\textit{offsetx}: the horizontal offset of polygons & \textit{offsetx}: the horizontal offset of circles \\ 
\textit{offsety}: the vertical offset of polygons & \textit{offsety}: the vertical offset of circles \\ \hline
\multicolumn{2}{|c|}{\textbf{Unchangeable Concept Set}} \\
\hline
\textit{rotation}: the rotation of inner polygons & $\varnothing$ \\ \hline
\end{tabular}
\vspace{0.2cm}
\caption{Detailed illustrations about the RPM-like dataset.}
\label{tab:detail_dataset}
\end{table*}

Table \ref{tab:detail_dataset} and Figure \ref{fig:sup_dataset} display the definition and samples of various instances on our RPM-like datasets, including the changeable set, unchangeable set and rule set. Definitions of rules in the rule set are:
\begin{itemize}
	\item \textbf{progress}: one concept will change with isometric steps in the horizontal or vertical direction (e.g., 2nd sample in Figure \ref{fig:sup_dataset}(a));
	\item \textbf{biprogress}: one concept will change with isometric steps in \textit{both} directions (e.g., 1st sample in Figure \ref{fig:sup_dataset}(a));
	\item \textbf{multiprogress}: \textit{multiple concepts} will change with isometric steps in the horizontal or vertical direction (e.g., 5th sample in Figure \ref{fig:sup_dataset}(b)).
\end{itemize}
Since we have described the Polygon and Circle dataset, in this section we only explain cell images inside the Complex Polygon, Complex Circle, Position Polygon and Position Circle dataset.
\begin{itemize}
\item \textbf{Complex Polygon.} A cell image is composed of a centered inner polygon and outer polygon. For the outer polygon, rotation and color are fixed, leaving size to be changeable. Inner polygons vary in the color and rotation while keeping the size unchangeable.

\item \textbf{Complex Circle.} On the basis of the Circle dataset, a centered inner circle is laid on the original cell image to form a more complex cell image. In addition to the concept-changing rules in the Circle dataset, new rules can change the grayscale of inner circles.

\item \textbf{Position Polygon} and \textbf{Position Circle}. In these instances of RPM-like datasets, objects are not centered and can be moved horizontally or vertically, which are described by the changeable concepts \textit{offsetx} and \textit{offsety}. The rules are then applied in the  concepts \textit{offsetx} and \textit{offsety}. 
\end{itemize}
\section{Hyperparameter Choice and Model Architecture}

\begin{table*}[t]
\centering
\begin{tabular}{|l|l|l|}
\hline
\textbf{Content Encoder $g_{\boldsymbol{\phi_{z}}}$} & \multicolumn{2}{l|}{\textbf{Location Encoder $g_{\boldsymbol{\phi_{\lambda}}}$}} \\ \hline
\multicolumn{3}{|l|}{Input 64 $\times$ 64 grayscale images} \\
\hline
4 $\times$ 4 conv. 32 BN. ReLU. stride 2. padding 1 & \multicolumn{2}{l|}{4 $\times$ 4 conv. 32 BN. ReLU. stride 2. padding 1} \\ \hline
4 $\times$ 4 conv. 32 BN. ReLU. stride 2. padding 1 & \multicolumn{2}{l|}{4 $\times$ 4 conv. 32 BN. ReLU. stride 2. padding 1} \\ \hline
4 $\times$ 4 conv. 64 BN. ReLU. stride 2. padding 1 & \multicolumn{2}{l|}{4 $\times$ 4 conv. 32 BN. ReLU. stride 2. padding 1} \\ \hline
4 $\times$ 4 conv. 128 BN. ReLU. stride 2. padding 1 & \multicolumn{2}{l|}{4 $\times$ 4 conv. 32 BN. ReLU. stride 2. padding 1} \\ \hline
4 $\times$ 4 conv. 256 BN. ReLU. & \multicolumn{2}{l|}{4 $\times$ 4 conv. 32 BN. ReLU.} \\ \hline
1 $\times$ 1 conv. $2 \times d_{\boldsymbol{z}}$. & FC. 128 ReLU. & FC. 128 ReLU. \\ \hline
& FC. $6 \times d_{\boldsymbol{z}} \times d_{loc}$. & FC. $6 \times d_{\boldsymbol{z}} \times d_{loc}$. \\ \hline
\end{tabular}
\vspace{0.2cm}
\caption{Architecture of the content encoder and location encoder where conv $=$ convolutional layer, BN $=$ Batch Normalization, ReLU $=$ Rectified Linear Unit, and FC $=$ Fully Connected layer. Outputs of the $g_{\boldsymbol{\phi_{z}}}$ are means and logarithmic variances of the latent variable $\boldsymbol{z}$ and we denote $\boldsymbol{z}$'s dimension as $d_{\boldsymbol{z}}$. In the location encoder, two branches of fully connected layers respectively represent the row and column latent variables. To represent dimensions with row and column latent variables, output dimension of $g_{\boldsymbol{\phi_{\lambda}}}$ is multiplied by $d_{\boldsymbol{z}}$.}
\label{tab:content_location_encoder}
\end{table*}

\begin{table*}[t]
\centering
\begin{tabular}{|l|l|}
\hline
\textbf{Kernel Encoder $g_{\boldsymbol{\theta}}$} & \textbf{Decoder $g_{\boldsymbol{\varphi}}$} \\ \hline
Input 2 $\times$ $d_{axis}$ latent variables & Input $d_{\boldsymbol{z}}$ latent variables \\
\hline
FC. 32 ReLU. & 1 $\times$ 1 deconv. 256 BN. ReLU. stride 1. \\ \hline
FC. 64 ReLU. & 4 $\times$ 4 deconv. 64 BN. ReLU. stride 1. \\ \hline
FC. 64 ReLU. & 4 $\times$ 4 deconv. 32 BN. ReLU. stride 2. padding 1 \\ \hline
FC. 64 ReLU. & 4 $\times$ 4 deconv. 32 BN. ReLU. stride 2. padding 1 \\ \hline
FC. 8 Tanh. & 4 $\times$ 4 deconv. 32 BN. ReLU. stride 2. padding 1 \\ \hline
& 4 $\times$ 4 deconv. 1 Sigmoid. stride 2. padding 1\ \\ \hline
\end{tabular}
\vspace{0.2cm}
\caption{Architecture of the kernel encoder and decoder where deconv $=$ deconvolutional layer. The kernel encoder $g_{\boldsymbol{\theta}}$ maps 2D locations $\boldsymbol{\lambda}$ to 8-dimension vectors for the RBF kernel. The decoder $g_{\boldsymbol{\varphi}}$ is utilized to construct cell images from latent variables.}
\label{tab:kernel_encoder_decoder}
\end{table*}

\begin{table*}[t]
\centering
\begin{tabular}{ |c|c|c|c|c| }
\hline
& Polygon & Circle & Complex Polygon \& Circle & Position Polygon \& Circle \\ \hline
$\beta$ & 30 & 20 & 50 & 30 \\ \hline
$\gamma$ & 30 & 20 & 50 & 30 \\ \hline
$d_{\boldsymbol{z}}$ & 5 & 5 & 10 & 10 \\ \hline
\end{tabular}
\vspace{0.2cm}
\caption{Hyperparameter values for different datasets.}
\label{tab:hyper}
\end{table*}

For all datasets, the proposed model uses the same architecture of encoders and decoders as Table \ref{tab:content_location_encoder} and Table \ref{tab:kernel_encoder_decoder} show. When training on various datasets, the hyperparameter learning rate is $1 \times 10^{-4}$, batch size is 256, and $d_{loc}=4$. Other hyperparameters are distinct among datasets whose values are displayed in Table \ref{tab:hyper}. In terms of model selection, all models are trained with the Adam algorithm \cite{kingma2014adam}. During the training process, models acquiring the highest ELBO on the validation set are selected as final models.\footnote{Code is available at https://github.com/FudanVI/generative-abstract-reasoning/tree/main/rpm-lgpp}

\section{Experiment Results On Additional Datasets}

\subsection{Quantitative Results}

\begin{figure*}[t]
  \centering
  \includegraphics[width=0.9\textwidth]{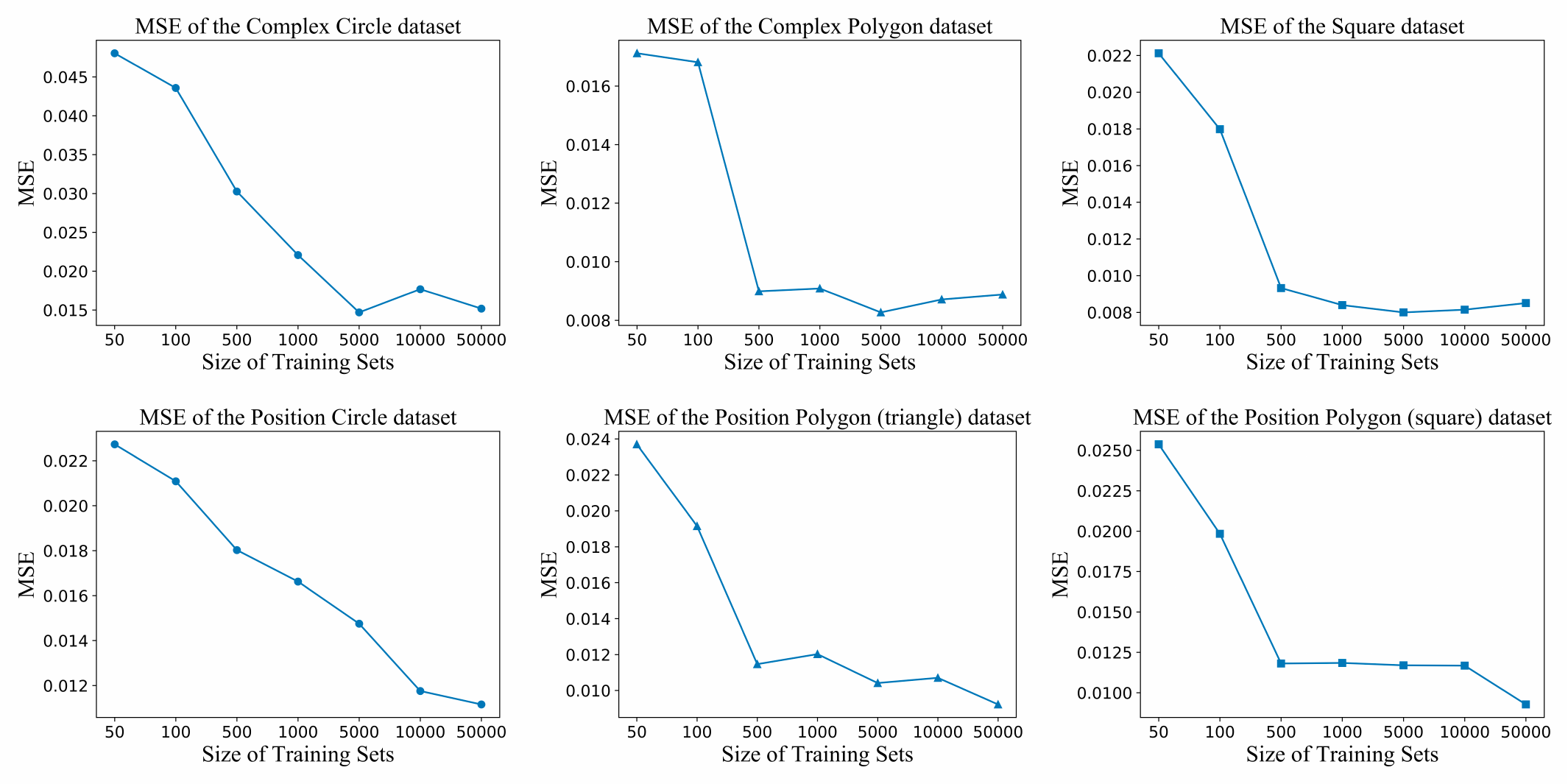}
  \caption{MSEs of additional datasets with the increasing size.}
  \label{fig:sup_mse}
\end{figure*}

\begin{table*}[t]
\centering
\begin{tabular}{ |c|c|c|c| }
\hline
& Square & Complex Polygon & Complex Circle \\ \hline
VAE-GAN & 4.31e-2 & 3.47e-2 & 1.98e-1 \\ \hline
Proposed & \textbf{8.00e-3 $\pm$ 2.6e-5} & \textbf{8.27e-3 $\pm$ 3.1e-5} & \textbf{1.47e-2 $\pm$ 5.2e-5} \\ \hline \hline

& Position Polygon (triangle) & Position Polygon (square) & Position Circle \\ \hline
VAE-GAN & 2.40e-2 & 3.55e-2 & 1.68e-2 \\ \hline
Proposed & \textbf{9.22e-3 $\pm$ 1.7e-5} & \textbf{9.28e-3 $\pm$ 2.1e-5} & \textbf{1.12e-2 $\pm$ 1.7e-5} \\ \hline
\end{tabular}
\vspace{0.2cm}
\caption{MSEs for the VAE-GAN and our model.}
\label{tab:sup_mse}
\end{table*}

We describe the data efficiency of our model on additional datasets in Figure \ref{fig:sup_mse}. For the Complex Polygon and Square dataset, our model achieves satisfying results with only 500 samples. For the Complex Circle dataset, our model can achieve well performance with 5000 samples. For the Position Polygon and Position Circle datasets, we set the number of samples to 50000. By comparing with the VAE-GAN, Table \ref{tab:sup_mse} displays outstanding MSE scores of our model, indicating that the cell prediction ability can be generalized to various instances of the RPM-like dataset.

\subsection{Cell Prediction}

\begin{figure*}[t]
  \centering
  \includegraphics[width=0.7\textwidth]{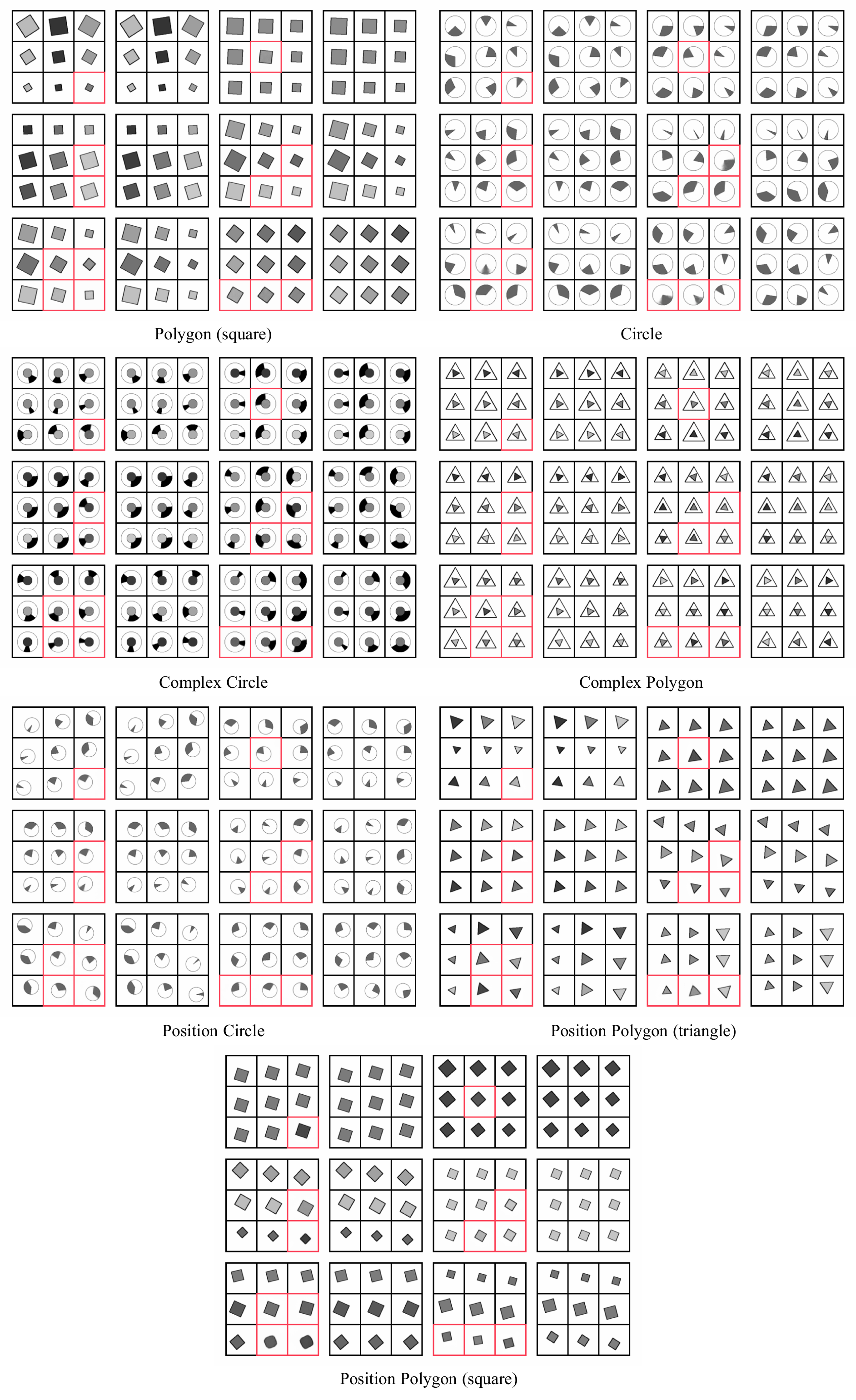}
  \caption{Cell prediction results of our model on additional datasets. Red-bordered cells contain model predictions and remaining cell images are ground truths.}
  \label{fig:sup_cell_prediction}
\end{figure*}

\begin{figure*}[t]
  \centering
  \includegraphics[width=0.8\textwidth]{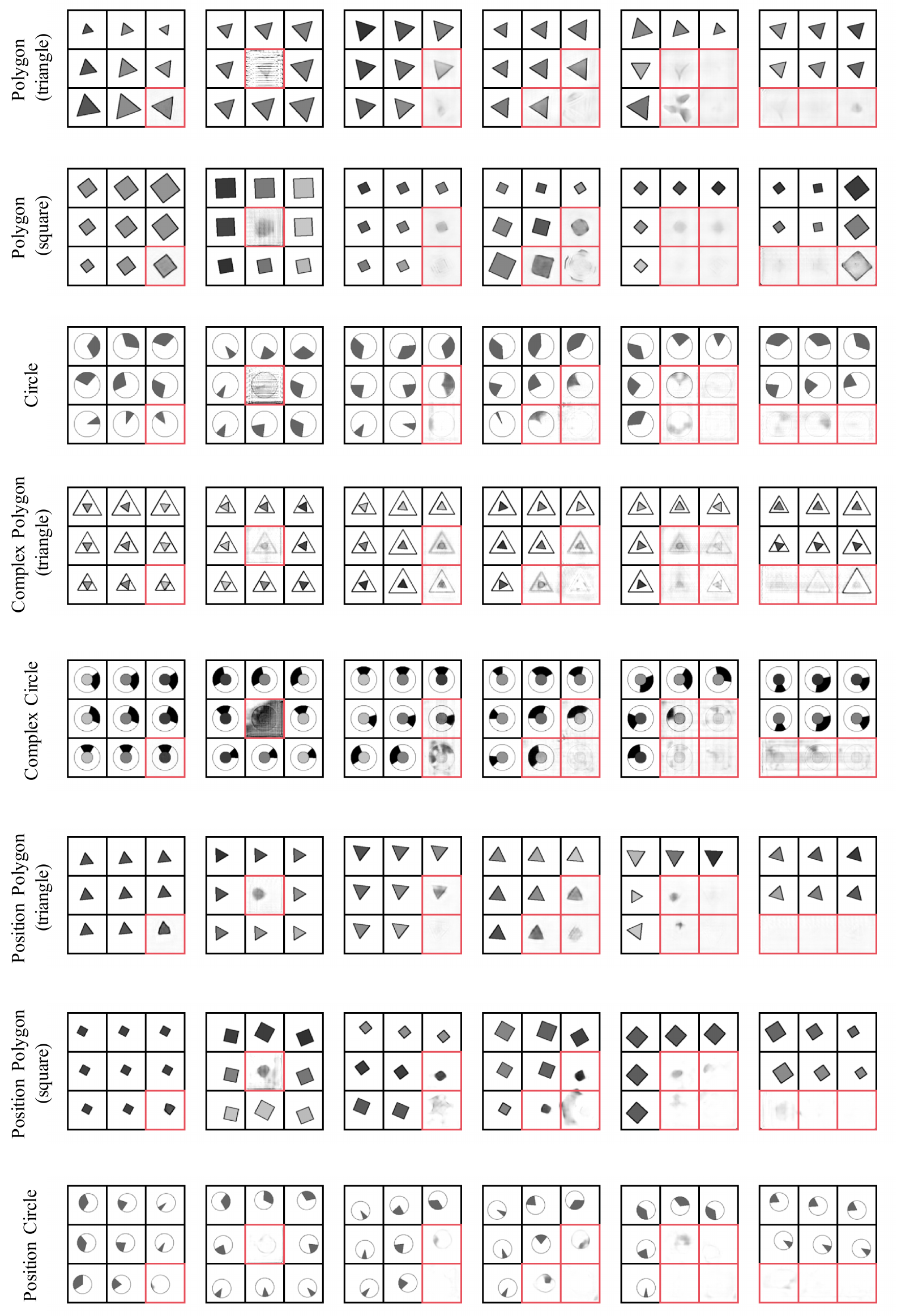}
  \caption{Cell prediction results of VAE-GAN on additional datasets. Red-bordered cells contain model predictions and remaining cell images are ground truths.}
  \label{fig:sup_cell_prediction_vaegan}
\end{figure*}

In Figure \ref{fig:sup_cell_prediction} and \ref{fig:sup_cell_prediction_vaegan}, we visualize the cell prediction results of our model and VAE-GAN on the additional datasets. Our model can perform the multi-cell and changing-position prediction in additional datasets as well, and the deviation between predictions and ground truths occurs in the concept \textit{grayscale} when fewer panel cells are given. In spite of the deviation, the model succeeds to capture concept-changing rules within context panels on the additional datasets to generate satisfying results. In contrast, VAE-GAN can only generate high-quality predictions for the right-bottom panel cells. Once the position or number of target panel cells changes, VAE-GAN can hardly generate clear predictions. These results show that VAE-GAN does not completely understand concept-changing rules in panels, and tries to directly learn a mapping from context cells to target cells, which lacks the generalization ability for the multi-cell and changing-position prediction.

\subsection{Panel Sampling and Interpolation}

\begin{figure*}[t]
  \centering
  \includegraphics[width=0.9\textwidth]{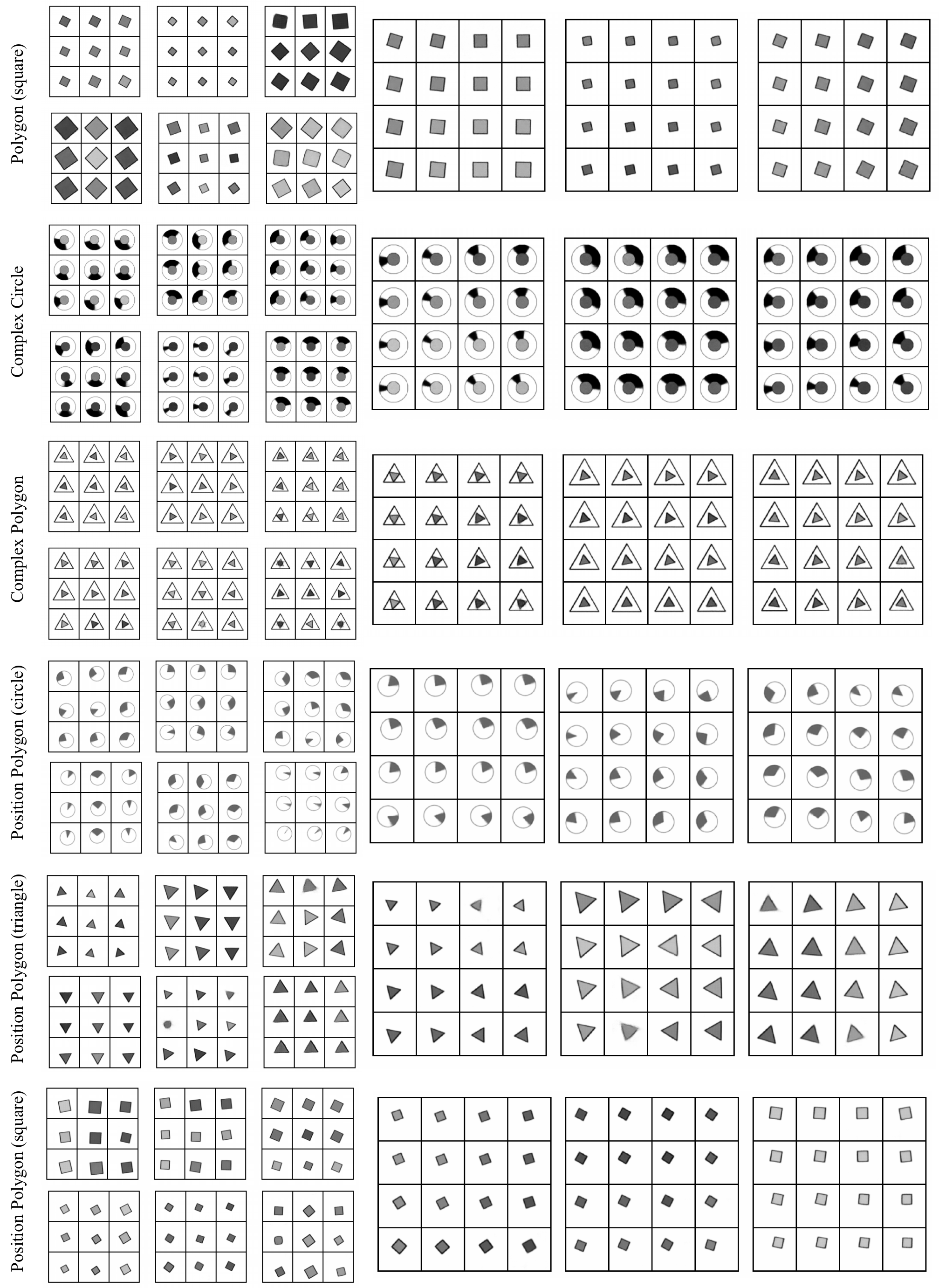}
  \caption{Generated (left) and interpolated (right) panels from our model.}
  \label{fig:sup_generation}
\end{figure*}

Detailed experiment results about the additional datasets are exhibited in Figure \ref{fig:sup_generation}. Results of the Polygon and Position datasets indicate that our model's learnable latent GP priors are adaptive, enabling us to generate panels within the predefined rule set. The sampling diversity of the Complex Polygon and Complex Circle dataset is a little inferior, because of the difficulty for independent complex visual concept inference. Most generated panels of the additional datasets still have meaningful concept-changing modes, conforming to the rule \textit{progress}.

\subsection{Interpretable Visual Concepts}

\begin{figure*}[t]
  \centering
  \includegraphics[width=0.9\textwidth]{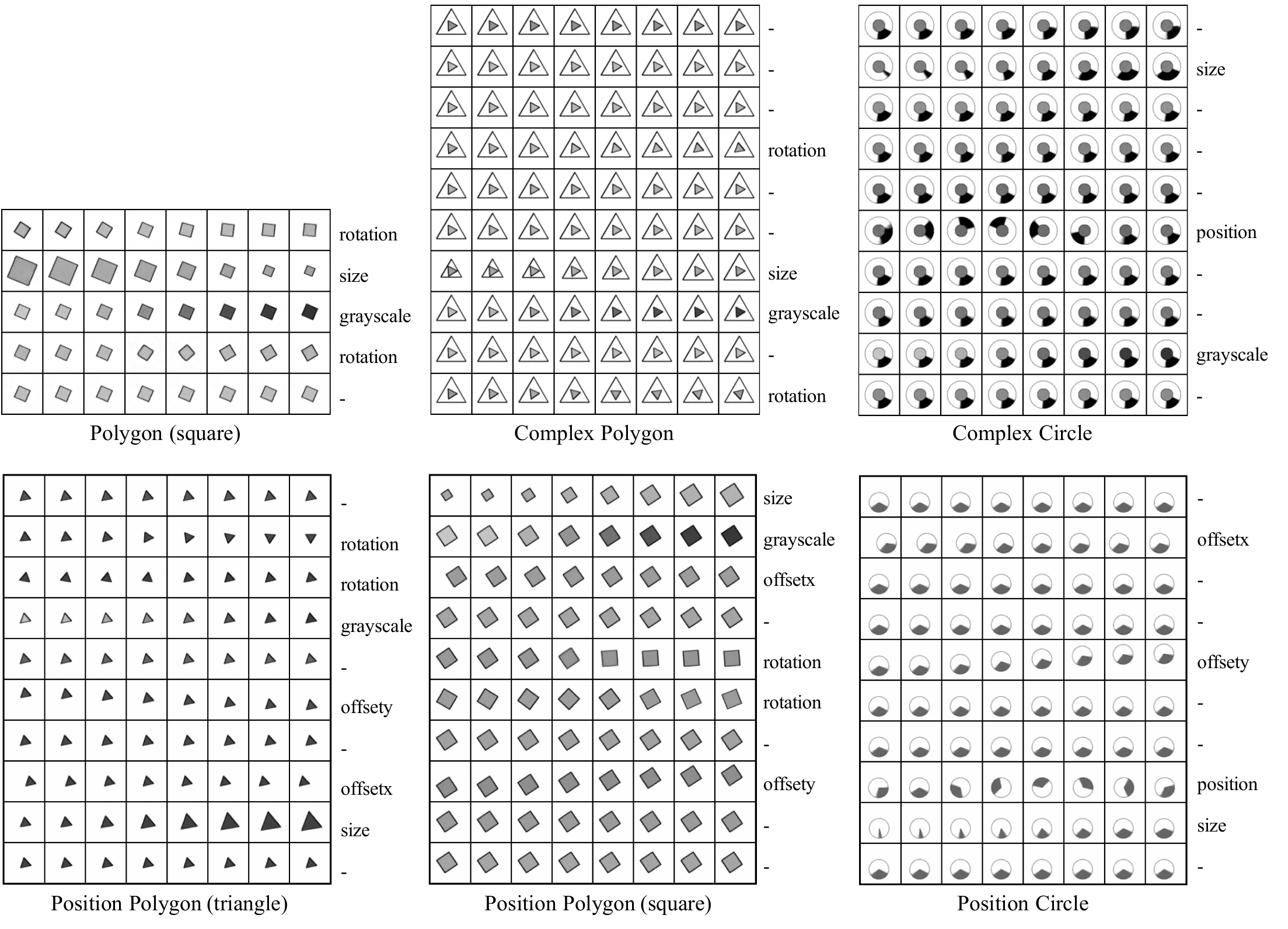}
  \caption{Dimensional cell image latent variable traversals on additional datasets. On the right side of each subfigure are related visual concepts of each dimension.}
  \label{fig:sup_disentangle}
\end{figure*}

\begin{table*}[t]
\centering
\begin{tabular}{ |c|c|c|c| }
\hline
\multicolumn{4}{|c|}{Factor VAE Score} \\ \hline
& Square & Complex Polygon & Complex Circle \\ \hline
$\beta$-VAE & 9.99e-1 $\pm$ 6.3e-4 & \textbf{1.00 $\pm$ 0.0} & 8.69e-1 $\pm$ 2.3e-2 \\ \hline
Proposed & \textbf{1.00 $\pm$ 0.0} & \textbf{1.00 $\pm$ 0.0} & \textbf{1.00 $\pm$ 0.0} \\ \hline

& Position Polygon (triangle) & Position Polygon (square) & Position Circle \\ \hline
$\beta$-VAE &  8.69e-1 $\pm$ 1.7e-2 & 8.84e-1 $\pm$ 1.6e-2 & 7.45e-1 $\pm$ 2.5e-2 \\ \hline
Proposed & \textbf{1.00 $\pm$ 0.0} & \textbf{9.99e-1 $\pm$ 1.4e-3} & \textbf{1.00 $\pm$ 0.0} \\ 

\hline \hline

\multicolumn{4}{|c|}{SAP Score} \\ \hline
& Square & Complex Polygon & Complex Circle \\ \hline
$\beta$-VAE & 4.28e-1 $\pm$ 2.6e-3 & 6.57e-1 $\pm$ 3.0e-3 & 1.56e-1 $\pm$ 2.9e-3 \\ \hline
Proposed & \textbf{8.21e-1 $\pm$ 2.7e-3} & \textbf{6.76e-1 $\pm$ 2.1e-3} & \textbf{9.07e-1 $\pm$ 9.9e-4} \\ \hline

& Position Polygon (triangle) & Position Polygon (square) & Position Circle \\ \hline
$\beta$-VAE & 3.69e-1 $\pm$ 2.7e-3 & 3.68e-1 $\pm$ 2.3e-3 & 5.12e-2 $\pm$ 3.3e-3 \\ \hline
Proposed & \textbf{8.92e-1 $\pm$ 1.5e-3} & \textbf{8.68e-1 $\pm$ 1.6e-3} & \textbf{9.23e-1 $\pm$ 1.6e-3} \\ \hline

\end{tabular}
\vspace{0.2cm}
\caption{Factor VAE and SAP Scores for the $\beta$-VAE and our model.}
\label{tab:sup_disentangle}
\end{table*}

For the Factor VAE Score, we sample 1000 training data points and each of which has 512 cell images with one unchanged concept. To obtain the SAP Score, $1 \times 10^{4}$ representation-concept pairs are generated. Considering complex cell images to have more concepts, we set $d_{\boldsymbol{z}}=10$ on the Complex and Position datasets while $d_{\boldsymbol{z}}=5$ on other datasets. Figure \ref{fig:sup_disentangle} suggests that when the dimension of latent variables is adequate for concept inference on datasets, our model tends to deduce interpretable visual concepts and leave other dimensions useless by inductive biases from the latent GP priors. Scores in Table \ref{tab:sup_disentangle} reveal that our model outperforms the basic $\beta$-VAE on additional RPM-like datasets as well.

\subsection{Selection Tasks}

\begin{table*}[t]
\centering
\begin{tabular}{ |c|c|c|c| }
\hline
& Proposed & WReN & CoPINet \\ \hline
Polygon (triangle) & 8.48e-1 $\pm$ 3.6e-3 & \textbf{9.99e-1} & 9.61e-1 \\ \hline
Polygon (square) & 8.49e-1 $\pm$ 2.1e-3 & \textbf{9.98e-1} & 9.75e-1 \\ \hline
Circle & 8.45e-1 $\pm$ 1.9e-3 & \textbf{9.99e-1} & 9.62e-1 \\ \hline
Complex Polygon (triangle) & 5.55e-1 $\pm$ 1.9e-3 & \textbf{9.99e-1} & 9.84e-1 \\ \hline
Complex Circle & 7.15e-1 $\pm$ 3.3e-3 & \textbf{9.99e-1} & 9.63e-1 \\ \hline
Position Polygon (triangle) & 8.01e-1 $\pm$ 2.6e-3 & \textbf{9.97e-1} & 9.61e-1 \\ \hline
Position Polygon (square) & 8.48e-1 $\pm$ 3.6e-3 & \textbf{9.95e-1} & 9.58e-1 \\ \hline
Position Circle & 7.65e-1 $\pm$ 3.3e-3 & \textbf{9.89e-1} & 9.31e-1 \\ \hline

\end{tabular}
\vspace{0.2cm}
\caption{Selection accuracy of our and compared models.}
\label{tab:acc}
\end{table*}

To evaluate the performance of our model in selection tasks, we create selective versions of the RPM-like datasets by introducing 8-cell selection panels. To endow the proposed model with the selective ability, we first calculate MSE scores between the prediction and 8 selections, then the one with the lowest MSE score is chosen as the answer. In the experiment, we compare the proposed model with WReN \cite{barrett2018measuring} and CoPINet \cite{zhang2019learning}. The selection accuracy is displayed in Table \ref{tab:acc} where WReN outperforms other models. Since our model focuses on the generative task, the noise in generations will influence the distinguishability of MSE scores, which leads to a relatively lower selection accuracy. On the other side, both compared models are designed for selective tasks, which is the objective of this experiment, they are reasonable to get higher performances. 

\end{document}